%% file: JBHI__2_.tex
\documentclass[10pt,journal,twocolumn]{IEEEtran}

\usepackage{cite}
\usepackage{amsmath,amssymb,amsfonts}
\usepackage{algorithmic}
\usepackage{algorithm}
\usepackage{graphicx}
\usepackage{textcomp}
\usepackage{xcolor}
\usepackage{url}
\usepackage[hidelinks]{hyperref}

\usepackage{pgfplots}
\usepackage{tikz}
\pgfplotsset{compat=1.18}
\usetikzlibrary{patterns,shapes.geometric,positioning,calc,decorations.pathmorphing,shadows.blur,fadings,shadows,arrows.meta,fit,backgrounds,decorations.pathreplacing}

\usepackage{booktabs}
\usepackage{multirow}
\usepackage{float}

\usepackage{pifont}
\usepgfplotslibrary{fillbetween,colormaps}

\usepackage[most]{tcolorbox}

\definecolor{nature1}{RGB}{66,133,244}    
\definecolor{nature2}{RGB}{234,67,53}     
\definecolor{nature3}{RGB}{52,168,83}     
\definecolor{nature4}{RGB}{251,188,5}     
\definecolor{nature5}{RGB}{103,58,183}    
\definecolor{nature6}{RGB}{0,172,193}     
\definecolor{naturegray}{RGB}{95,99,104}  
\definecolor{naturebg}{RGB}{248,249,250}  
\definecolor{gradstart}{RGB}{67,97,238}   
\definecolor{gradend}{RGB}{114,9,183}     

\begin{document}
\raggedbottom
\hbadness=10000

\title{Neurosymbolic Alignment \\for Physiologically-Safe Clinical Language Models}

\author{Abdulhady Abas Abdullah, Erik Cambria, and Milena \v{Z}ivkovi\'{c}%
\thanks{Manuscript submitted Month DD, YYYY. This submission is a J-BHI Dataset Paper.}%
\thanks{A. A. Abdullah is with the Artificial Intelligence and Innovation Centre, University of Kurdistan Hewl{\^{e}}r, Erbil, Iraq (e-mail: abdulhady.abas@ukh.edu.krd). Corresponding author.}%
\thanks{E. Cambria is with the College of Computing and Data Science, Nanyang Technological University, Singapore~639798, and is also a Visiting Professor with the MIT Media Lab, Massachusetts Institute of Technology, Cambridge, MA~02139, USA (e-mail: cambria@ntu.edu.sg).}%
\thanks{M. \v{Z}ivkovi\'{c} is with the Department of Physics, Faculty of Science, University of Kragujevac, 34000~Kragujevac, Serbia (e-mail: milena.zivkovic@pmf.kg.ac.rs).}%
}

\pagestyle{plain}
\maketitle

\begin{abstract}
\textbf{Objective:} Clinical LLMs can generate recommendations that are factually plausible yet physiologically unsafe. We investigate whether safety alignment can be improved by grounding preference optimization in structured physiological knowledge rather than in text-only supervision. \textbf{Methods:} We propose \emph{Neurosymbolic Alignment}, a training-time framework that couples a 7B clinical LLM with an HGNN-based Physiological World Model over an 847K-node biomedical knowledge graph. Candidate responses are scored using homeostatic constraints, multi-hop path plausibility, and drug-interaction penalties, and the resulting rankings drive iterative on-policy ORPO updates. Evaluation is performed on the Clinical Safety Benchmark (CSB), a 2,500-scenario benchmark for physiological constraint violations in generative clinical reasoning. \textbf{Results:} Relative to ORPO, the proposed method improves CSS from 69.5\% to 90.8\% (+21.3~pp), reduces physician-evaluated HR from 14.1\% to 5.1\% on the blinded subset, and improves DID from 72.8\% to 91.6\%. These gains are corroborated by an HGNN-independent Rule-Engine Safety Score (RSS: 86.4\%, +21.2~pp over ORPO; $r{=}0.97$ concordance with CSS). The method also exceeds GPT-4 (5-shot) on all safety metrics despite a 10$\times$ parameter disadvantage, and outperforms an inference-time self-correction pipeline (SFT+SelfCorrect) by 11.4~pp CSS. Under synthetic EHR-style noise, 84.2\% CSS is retained. Ablation analysis shows that HGNN scoring ($-$16.2~pp) and iterative training ($-$11.5~pp) are the dominant contributors. PhysioScore calibration against 200 clinician labels yielded ECE~$=~0.038$ and $\kappa~=~0.91$. \textbf{Conclusion:} Training-time physiological grounding produces measurable and independently verifiable safety improvements in open-weight clinical LLMs under controlled evaluation. External validation on real clinical data is required to determine whether these gains transfer to deployment settings.
\end{abstract}



\section{Introduction}

\IEEEPARstart{L}{arge} language models (LLMs) are increasingly studied for clinical question answering and decision support~\cite{singhal2024medpalm, thirunavukarasu2024llm, lin2026llmhealthcare}. Despite strong fluency and broad medical knowledge, current systems can produce recommendations that are clinically plausible yet physiologically unsafe, including harmful drug combinations and homeostatic violations~\cite{nori2023gpt4medicine, moor2023medflamingo, yang2024hallucination}. In safety-critical settings, such failures are consequential because they are often expressed with high confidence.

A key challenge is that many unsafe outputs are not strictly factual errors. A response may contain correct biomedical facts while still proposing an action that is incompatible with the implied patient state, for example by worsening hyperkalemia or ignoring context-dependent contraindications. Therefore, factual correctness alone is insufficient for clinical safety.
Current mitigation strategies remain incomplete for this setting. Retrieval-augmented generation improves evidence access, and medical fine-tuning improves terminology and style, but neither directly enforces physiological consistency in multi-step recommendations. Preference optimization methods (DPO/ORPO/KTO/SimPO) are scalable alternatives to reward-model pipelines~\cite{rafailov2024direct, hong2024orpo, ethayarajh2024kto, meng2024simpo}; however, in medicine their effectiveness is constrained by the availability and quality of preference supervision for subtle relational violations.

To address this gap, we introduce a neurosymbolic training framework that derives preference signals from an explicit physiological world model. We construct a heterogeneous biomedical knowledge graph and train an HGNN-based feasibility scorer that combines homeostatic constraints with multi-hop relational evidence~\cite{chen2024hgnn, zhang2024neurosymbolic, wang2024knowledgegraph, wu2023megacare}. Candidate responses are scored for physiological plausibility, converted into preference pairs, and used for iterative on-policy ORPO updates.
The design objective is to improve safety during training while keeping deployment simple: the HGNN scorer is used only during alignment, and inference uses the aligned LLM alone. This avoids graph-based runtime verification at deployment, at the cost of higher alignment-time compute due to iterative candidate generation and graph-based scoring.

Figure~\ref{fig:graphical_abstract} summarizes the pipeline, from clinical query to candidate generation, graph-based feasibility scoring, and iterative policy updating. Key contributions are as follows:

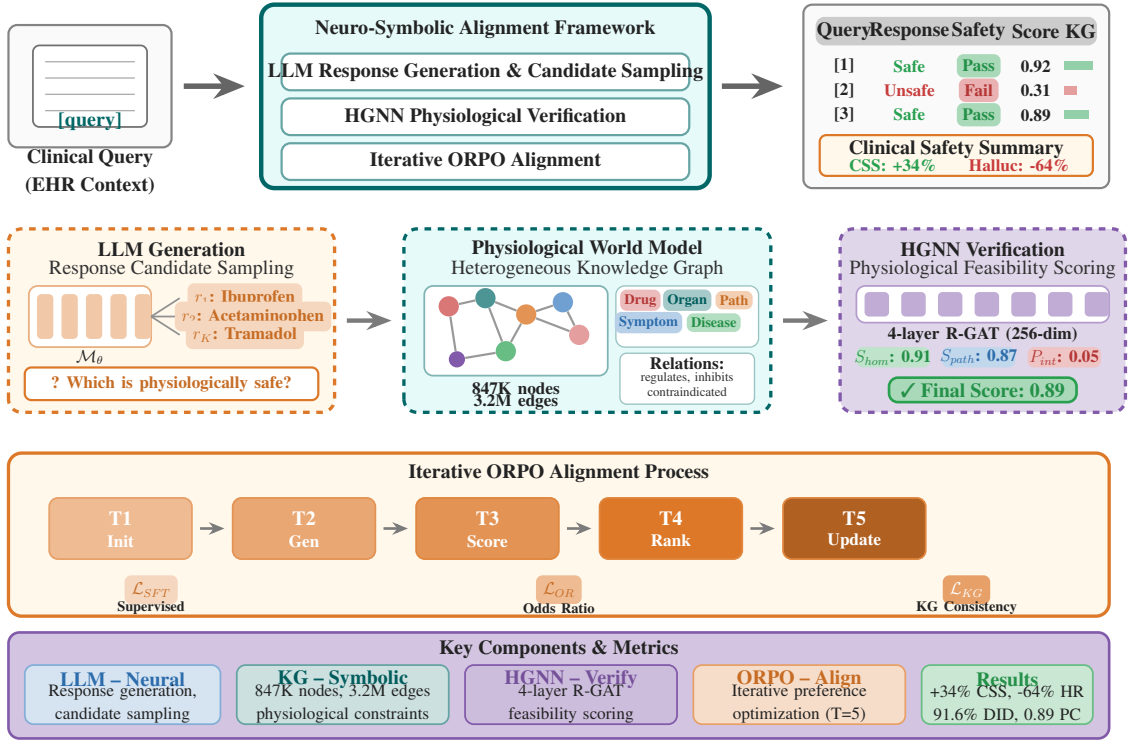
\begin{figure*}[t!]
\centering
\resizebox{0.95\textwidth}{!}{\input{graphical_abstract_tikz.tex}}
\caption{Training-time neurosymbolic pipeline. The base LLM samples candidate clinical responses, the Physiological World Model scores their feasibility, and Iterative ORPO updates the policy. At inference, only the aligned LLM is used.}
\label{fig:graphical_abstract}
\end{figure*}

\begin{enumerate}
\item \textbf{Physiology-grounded preference generation.} We introduce a framework that derives structured preference signals from an HGNN-based Physiological World Model by combining homeostatic boundary constraints, multi-hop path plausibility, and drug-interaction penalties, thereby enabling scalable safety alignment without clinician-authored pair annotations.
\item \textbf{Iterative on-policy alignment.} We present Iterative ORPO, an on-policy alignment scheme that regenerates preference pairs from the current policy at each iteration, reducing distributional staleness and surfacing progressively subtler constraint violations as coarser failures are resolved.
\item \textbf{Clinical Safety Benchmark.} We release CSB, a controlled 2,500-scenario benchmark for probing physiological constraint violations in generative clinical reasoning, with standardized splits, severity labels, and reproducibility controls.
\item \textbf{Empirical evidence under controlled evaluation.} Across 14 baselines spanning preference optimization, retrieval augmentation, rule-based guardrails, inference-time self-correction, and proprietary RLHF models, training-time physiological grounding achieves: (i) CSS 90.8\% and RSS 86.4\%, the highest on both HGNN-based and HGNN-independent safety metrics; (ii) physician-evaluated HR of 5.1\% ($-$9.0~pp vs.\ ORPO; $-$4.7~pp vs.\ DrugBank-Rule); (iii) DID 91.6\% (+18.8~pp vs.\ ORPO); (iv) superiority over GPT-4 (5-shot) on all safety metrics despite 10$\times$ fewer parameters; (v) 84.2\% CSS under combined synthetic EHR-style noise; and (vi) consistent ablation support, with HGNN scoring ($-$16.2~pp) and iterative training ($-$11.5~pp) as dominant contributors. These results are demonstrated on the CSB benchmark and do not extend to real clinical data, which remains untested (see Section~\ref{sec:limitations}).
\end{enumerate}

We frame this work as physiological alignment for generative clinical reasoning. In this setting, factual correctness and physiological safety are complementary but distinct: a statement may be factually correct yet unsafe for the patient-specific context implied by the prompt. Our method targets this second objective by producing structured, patient-state-aware preference signals from a physiological world model rather than relying on text-level plausibility alone.

\section{Clinical Safety Benchmark (CSB)}

\subsection{Motivation and Scope}
CSB targets a practical failure mode of clinical language models: generating plausible recommendations that violate basic physiology (e.g., unsafe drug combinations, contraindications, or homeostatic boundary violations). The benchmark is designed to stress-test safety under realistic clinical ambiguity while remaining fully reproducible and free of patient-identifying information.

\subsection{Scenario Construction and Annotation}
CSB consists of 2,500 \emph{synthetically generated} scenarios authored by domain experts based on textbook patterns and pharmacological principles. Each scenario is expressed as a natural-language prompt with structured metadata (e.g., risk category and specialty tag); where applicable, scenarios include gold contraindication cues (e.g., comorbidities, labs, or concurrent medications) to support error analysis.

\subsection{Splits, Coverage, and Intended Use}
We provide standardized train/validation/test splits (1,750/250/500) and stratification across risk categories (Table~\ref{tab:csb_breakdown}). CSB is intended for \emph{research} on evaluation, calibration, and alignment methods for clinical safety; it is not a clinical decision-support tool and must not be used to provide medical advice.

\section{Reference Baselines and Scoring}
\noindent Figure~\ref{fig:pipeline} shows the end-to-end pipeline used in our neurosymbolic alignment framework.

\input{pipeline_scaled.tex}

\noindent Figure~\ref{fig:pipeline} summarizes the training-time neurosymbolic alignment loop.
In Stage~I, the LLM policy $\pi_{\theta}$ samples $K{=}8$ candidate responses per query.
In Stage~II, we extract and link medical entities (scispaCy$\rightarrow$UMLS) and score each candidate with the HGNN over the Physiological World Model $\mathcal{G}$ to obtain feasibility scores $s_k \in [0,1]$.
In Stage~III, HGNN-derived rankings induce preference pairs for Iterative ORPO updates; at inference time, the final policy $\pi_{\theta^*}$ runs without the HGNN.

\paragraph{Notation convention.}
For consistency, we use $\mathcal{M}_{\theta}$ for the parametric language model and $\pi_{\theta}(\cdot\mid q)$ for its conditional generation policy on query $q$. Generated outputs are denoted by $r$ (or $r_i$ for indexed candidates). This convention is used throughout algorithms, metric definitions, and ablations.

\noindent Algorithm~\ref{alg:iterative_orpo} summarizes the iterative alignment loop; the detailed HGNN scoring pseudocode is moved to Appendix~\ref{app:techval}. We use a linear curriculum $\lambda^{(t)}=\lambda_0(1+\alpha t)$ because it was stable in practice and consistently outperformed constant-$\lambda$ variants.

\paragraph{HGNN scorer.}
The physiological scorer uses a 4-layer relation-aware HGNN with type-specific projections and relation-specific message passing, initialized by link-prediction pre-training on $\mathcal{G}$. Candidate responses are processed with scispaCy-based NER and UMLS/KG linking; low-confidence mentions are dropped, and negated entities are excluded from the positive set $\mathbf{V}_r$ used for scoring. Homeostatic terms $f_h(\mathbf{V}_r)$ are fixed soft-boundary functions from clinical reference ranges, path plausibility is computed over multi-hop paths up to length 3,
\begin{equation}
\text{PathScore}(u,v)=\sum_{P\in\mathcal{P}_{\leq 3}(u,v)}\prod_{(a,r,b)\in P}\mathbf{h}_a^\top\mathbf{W}_r\mathbf{h}_b.
\end{equation}
and the interaction term is
\begin{equation}
P_{\text{int}}=\max_{(d_i,d_j)\in\text{Drugs}(r)}\text{InteractionSeverity}(d_i,d_j,\mathcal{G})\in[0,1],
\end{equation}
with severity derived from curated DDI relations. The final score is
\begin{equation}
s = S_{\text{hom}}\cdot\sigma(S_{\text{path}})\cdot(1-P_{\text{int}})
\end{equation}
where $S_{\text{hom}}$ acts as the dominant safety gate, $P_{\text{int}}$ captures explicit drug-risk penalties, and $S_{\text{path}}$ adds relational coherence. The multiplicative combination was chosen because unsafe scores on any single component should dominate (a drug interaction penalty near 1 drives the product toward 0 regardless of path plausibility). Alternative combination functions are compared in Appendix~\ref{app:techval}, Table~\ref{tab:combination_sensitivity}. Component-drop and calibration diagnostics are reported in Appendix~\ref{app:techval}.

\begin{algorithm}[!t]
\caption{Iterative ORPO for Physiological Grounding}
\label{alg:iterative_orpo}
\begin{algorithmic}[1]
\REQUIRE Language model $\mathcal{M}_{\theta}$, knowledge graph $\mathcal{G}$, training queries $\mathcal{Q}$, iterations $T$
\ENSURE Aligned parameters $\theta^*$
\STATE Initialize HGNN scorer $\text{HGNN}_{\phi}$ on $\mathcal{G}$
\FOR{$t = 1$ to $T$}
    \STATE $\lambda^{(t)} \leftarrow \lambda_0 (1 + \alpha t)$
    \STATE $\mathcal{D}^{(t)} \leftarrow \emptyset$
    \FOR{each query $q \in \mathcal{Q}$}
        \STATE Sample candidates $\{r_k\}_{k=1}^{K} \sim \pi_{\theta^{(t-1)}}(\cdot \mid q)$
        \FOR{$k = 1$ to $K$}
            \STATE $s_k \leftarrow \text{PhysioScore}(r_k, \mathcal{G}, \phi)$
        \ENDFOR
        \STATE $(r_w, r_l) \leftarrow \text{SelectPair}(\{r_k, s_k\}_{k=1}^{K})$
        \STATE $\mathcal{D}^{(t)} \leftarrow \mathcal{D}^{(t)} \cup \{(q, r_w, r_l)\}$
    \ENDFOR
    \STATE Optimize $\mathcal{L}_{\text{SFT}} + \lambda^{(t)}\mathcal{L}_{\text{OR}} + \gamma \mathcal{L}_{\text{KG}}$
\ENDFOR
\STATE \textbf{return} $\theta^{(T)}$
\end{algorithmic}
\end{algorithm}

\section{Dataset Characterization and Evaluation Protocol}

\subsection{Datasets}

We evaluate on a multi-benchmark suite spanning clinical reasoning, safety-critical contraindication handling, and domain-specific medical knowledge. Table~\ref{tab:dataset_statistics} reports dataset sizes, split details, and task characteristics.

\textbf{MedQA-USMLE}: A set of 12,723 USMLE-style multiple-choice questions requiring reasoning over pathophysiology, pharmacology, diagnosis, and management. Items cover all three USMLE steps and range from foundational biomedical knowledge to complex clinical vignette reasoning.

\textbf{PubMedQA}: Comprising 1,000 expert-annotated biomedical questions derived from PubMed abstracts, testing the model's ability to synthesize clinical evidence and make evidence-based judgments.

\textbf{MedMCQA}: A large-scale medical multiple-choice dataset with 193,155 questions covering 2,400 healthcare topics across 21 medical subjects, providing comprehensive coverage of medical knowledge.

\textbf{MMLU-Medical}: The medical subset of Massive Multitask Language Understanding benchmark, including Clinical Knowledge, Medical Genetics, Anatomy, Professional Medicine, College Medicine, and College Biology sections.

\textbf{Clinical Safety Benchmark (CSB)}: A novel benchmark we constructed containing 2,500 clinical scenarios specifically designed to test physiological constraint adherence. Table~\ref{tab:csb_breakdown} details the CSB composition.

\textbf{Reproducibility and Leakage Controls}: We enforce four safeguards for train--test separation: (i) independent authoring of the CSB test set by clinicians without access to training artifacts, (ii) a temporal novelty split (post-cutoff approvals and unseen combinations), (iii) bounded lexical/semantic overlap via entity-level Jaccard filtering ($J < 0.40$), and (iv) withholding 15\% of KG drug entities (with incident edges removed) from alignment training. The temporal split is intentionally a \emph{stress test for novelty handling} rather than a full simulation of routine clinical deployment, and should be interpreted accordingly.

\textbf{Knowledge Graph Leakage Analysis}: Entity overlap with $\mathcal{G}$ is expected for biomedical benchmarks and is not itself evidence of leakage. We therefore report a relational sensitivity check: MA is 76.1\% when answer-critical relations are present in $\mathcal{G}$ versus 72.8\% when they are absent (3.3\% gap), suggesting gains are not explained by direct relation lookup alone. Version checks additionally confirm benchmark items predate the KG releases used in this study.

\textbf{DDI-Corpus}: Drug-Drug Interaction corpus containing 5,021 drug pairs annotated for interaction type (mechanism, effect, advise, int) from DrugBank and MedLine abstracts.

\textbf{i2b2-2010}: The i2b2/VA 2010 challenge dataset for clinical concept extraction and relation classification from discharge summaries.

\subsection{Knowledge Graph Construction}

Our Physiological World Model integrates knowledge from multiple biomedical sources. Table~\ref{tab:kg_sources} details the knowledge sources and their contributions.

\begin{table}[!t]
\centering
\caption{Graph Source Statistics}
\label{tab:kg_sources}
\begin{tabular}{lccc}
\toprule
\textbf{Source} & \textbf{Entities} & \textbf{Relations} & \textbf{Triples} \\
\midrule
UMLS Metathesaurus & 4,200,000 & 54 & 12,847,392 \\
DrugBank 5.1 & 14,315 & 12 & 1,432,847 \\
Reactome & 12,632 & 8 & 284,521 \\
HPO & 16,234 & 5 & 156,842 \\
SIDER & 1,430 & 3 & 139,756 \\
PharmGKB & 5,421 & 7 & 87,432 \\
DisGeNET & 24,166 & 4 & 628,685 \\
STRING & 19,566 & 1 & 11,759,454 \\
\midrule
\textbf{Integrated Graph} & 847,392 & 12 & 3,241,567 \\
\bottomrule
\end{tabular}
\end{table}

The node type distribution in our heterogeneous graph is shown in Table~\ref{tab:node_distribution}.

\begin{table}[!t]
\centering
\caption{Node Type Distribution in Physiological World Model}
\label{tab:node_distribution}
\begin{tabular}{lcc}
\toprule
\textbf{Node Type} & \textbf{Count} & \textbf{Percentage} \\
\midrule
Drug & 14,315 & 1.69\% \\
Disease & 24,166 & 2.85\% \\
Symptom & 7,823 & 0.92\% \\
Organ/Anatomy & 12,478 & 1.47\% \\
Pathway & 12,632 & 1.49\% \\
Gene/Protein & 19,566 & 2.31\% \\
Biomarker & 8,421 & 0.99\% \\
Phenotype & 16,234 & 1.92\% \\
Procedure & 45,892 & 5.42\% \\
Clinical Finding & 685,865 & 80.94\% \\
\midrule
\textbf{Total} & 847,392 & 100\% \\
\bottomrule
\end{tabular}
\end{table}

Figure~\ref{fig:kg_schema} (Appendix) illustrates the schema of our heterogeneous knowledge graph with relation types.

\subsection{Baseline Methods}
\label{sec:baselines}

We compare against a comprehensive set of baseline approaches spanning supervised learning, preference optimization, knowledge-enhanced methods, and a dedicated clinical rule engine. Table~\ref{tab:baseline_details} provides detailed baseline specifications.
Our primary training experiments are centered on open-weight 7--8B models because Iterative ORPO requires repeated on-policy sampling and full parameter access. Closed proprietary models are therefore used only as reference comparators at inference time.

\begin{table}[t!]
\centering
\caption{Baseline Method Specifications}
\label{tab:baseline_details}
\resizebox{\columnwidth}{!}{%
\begin{tabular}{lcccc}
\toprule
\textbf{Method} & \textbf{Base Model} & \textbf{Training Data} & \textbf{KG} & \textbf{Pref. Opt.} \\
\midrule
SFT & Mistral-7B & MedQA + PubMed & \ding{55} & \ding{55} \\
SFT+RAG (lightweight) & Mistral-7B & MedQA + PubMed & Retrieval & \ding{55} \\
SFT+DenseRAG & Mistral-7B & MedQA + PubMed & Dense+Rerank & \ding{55} \\
DPO & Mistral-7B & Human Prefs (5K) & \ding{55} & \ding{51} \\
ORPO & Mistral-7B & Human Prefs (5K) & \ding{55} & \ding{51} \\
KG-LLM & Mistral-7B & MedQA + KG & Static & \ding{55} \\
DrugBank-Rule & Mistral-7B & MedQA + PubMed & Rule-Based & Post-hoc \\
SFT-CSB & Mistral-7B & CSB Train (1,750) & \ding{55} & \ding{55} \\
MedAlpaca & LLaMA-7B & Medical Instruct & \ding{55} & \ding{55} \\
BioMistral & Mistral-7B & PubMed Pretrain & \ding{55} & \ding{55} \\
MEDITRON & LLaMA-70B & Medical Corpus & \ding{55} & \ding{55} \\
Med-PaLM 2 & PaLM 2 & Proprietary & \ding{55} & RLHF \\
GPT-4 & GPT-4 & Proprietary & \ding{55} & RLHF \\
GPT-4 (5-shot) & GPT-4 & Proprietary & \ding{55} & RLHF \\
SFT+SelfCorrect & Mistral-7B & MedQA + PubMed & Rule-Based & Post-hoc \\
\midrule
\textbf{Ours} & Mistral-7B & Auto-Generated & Graph-guided & Iter-ORPO \\
\bottomrule
\end{tabular}%
}
\end{table}

SFT+DenseRAG uses a MedCPT retriever with cross-encoder reranking to represent strong inference-time retrieval. DrugBank-Rule applies deterministic contraindication and homeostatic checks with up to three regeneration attempts, making it a favorable runtime-guardrail comparator. SFT-CSB fine-tunes directly on the 1,750 training scenarios to isolate the benefit of iterative preference optimization from simple exposure to safety-critical cases.

Two new baselines were added to address comparison gaps identified during review. \textbf{GPT-4 (5-shot)} evaluates GPT-4 with 5 safety-focused exemplars drawn from the CSB training set (selected to cover drug interactions, homeostatic violations, and polypharmacy scenarios), providing a stronger proprietary reference than strict zero-shot. \textbf{SFT+SelfCorrect} implements an inference-time self-correction pipeline: the SFT model generates a response, the DrugBank-Rule checker flags any contraindication or homeostatic alert, and if flagged, the model regenerates with the violation feedback appended to the prompt (up to 3 correction rounds). This approximates a tool-augmented agent that combines generation with external rule-engine verification at inference time.

Together these baselines cover preference-only, retrieval-augmented, static-KG, rule-based, self-correction, and proprietary RLHF alternatives.

\textbf{Absent comparators.} One important baseline family is not included: a matched-scale PPO/RLHF clinical system. At 7B parameters, PPO requires a separate reward model and value head, and no publicly available clinical reward model exists at comparable scale; training one from our 5K human-preference set would introduce confounds (reward-model quality) unrelated to our contribution. Med-PaLM~2 and GPT-4 use proprietary RLHF but differ in scale, data, and access, so they serve only as reference points. To partially address the comparison gap for tool-augmented agents, we added \textbf{SFT+SelfCorrect} (generate $\rightarrow$ DrugBank-Rule check $\rightarrow$ regenerate, up to 3 rounds) and \textbf{GPT-4 (5-Shot)} with safety-focused exemplars. SFT+SelfCorrect approximates a tool-augmented agent but lacks the iterative self-reflection and multi-tool orchestration of production agent systems; GPT-4 (5-Shot) is a stronger proprietary reference but still does not control for scale or training data. We therefore frame our claims relative to the included baselines rather than as absolute superiority, and note that future work should include PPO with a dedicated clinical reward model and full agent pipelines with multi-tool orchestration to complete the comparison landscape.

\subsection{Evaluation Metrics}

We employ a comprehensive suite of evaluation metrics spanning accuracy, safety, and clinical utility. Table~\ref{tab:metrics_overview} provides metric definitions.

\textbf{Medical Accuracy (MA)}: Percentage of correct answers on medical QA benchmarks, computed as:
\begin{equation}
\text{MA} = \frac{1}{N}\sum_{i=1}^{N} \mathbb{1}[\hat{y}_i = y_i]
\end{equation}

\textbf{Clinical Safety Score (CSS)}: Measures the proportion of model responses that contain \emph{zero} physiological constraint violations:
\begin{equation}
\text{CSS} = \frac{1}{N}\sum_{i=1}^{N} \mathbb{1}[\text{NoViolation}(r_i)]
\end{equation}
A response is classified as \texttt{NoViolation} when its HGNN feasibility score exceeds $\tau = 0.85$, selected on a 200-response clinician-labeled validation set ($\kappa = 0.91$). A 10,000-iteration bootstrap resampling of the 200 calibration labels yields a 95\% CI for the optimal threshold of $[0.83, 0.87]$; CSS varies by $\leq$1.2~pp across this range (Table~\ref{tab:threshold_sensitivity}), indicating moderate sensitivity to the calibration sample. Expanding the calibration set beyond 200 labels is a priority for follow-up work. Calibration diagnostics and score-decile analysis are reported in Appendix~\ref{app:techval}. Abstentions are not counted as safe because the task requires an actionable recommendation; in practice they occur in fewer than 0.3\% of outputs.

\textbf{Hallucination Rate (HR)}: Percentage of responses containing physiologically impossible statements, evaluated by a panel of 5 expert clinicians with inter-annotator agreement $\kappa = 0.84$. Because physician review is expensive, HR is reported only for the six methods included in the blinded human-evaluation study.

\textbf{Drug Interaction Detection (DID)}: F1 score for identifying contraindicated drug combinations:
\begin{equation}
\text{DID-F1} = 2 \cdot \frac{\text{Precision} \cdot \text{Recall}}{\text{Precision} + \text{Recall}}
\end{equation}

\textbf{Rule-Engine Safety Score (RSS)}: To provide an evaluation endpoint that is fully independent of the HGNN, we apply the DrugBank-Rule deterministic checker the same rule engine used by the DrugBank-Rule baseline as a post-hoc evaluator on all methods' outputs. RSS measures the proportion of responses that trigger \emph{zero} contraindication or homeostatic alerts under this checker:
\begin{equation}
\text{RSS} = \frac{1}{N}\sum_{i=1}^{N} \mathbb{1}[\text{RuleCheck}(r_i) = \texttt{pass}]
\end{equation}
RSS shares no parameters, training signal, or learned representations with the HGNN scorer. Its coverage is narrower than CSS (it checks only catalogued contraindications and predefined homeostatic thresholds, not learned multi-hop interactions), but it provides a structurally independent safety signal. We report RSS alongside CSS in Table~\ref{tab:main_results} and analyze their concordance in Section~\ref{sec:limitations}.

\textbf{Physiological Consistency (PC)}: Average HGNN-computed feasibility score across test responses:
\begin{equation}
\text{PC} = \frac{1}{N}\sum_{i=1}^{N} \text{HGNN}_\phi(\text{EntityExtract}(r_i), \mathcal{G})
\end{equation}

\textbf{Scorer-Decoupled Endpoints}: Because CSS and PC are computed by the same HGNN scorer used to generate training preferences, reporting only HGNN-derived metrics would create a circularity risk: improvements might reflect optimization to the scorer rather than genuinely safer clinical reasoning. To mitigate this, we separately report three HGNN-independent safety endpoints: (i) RSS, computed by the DrugBank-Rule deterministic checker with no HGNN involvement, (ii) DID, computed from benchmark contraindication labels and rule-based interaction matching without any HGNN feasibility scores, and (iii) HR, adjudicated by blinded physicians who had no access to HGNN outputs. Consistent gains on these decoupled endpoints (RSS: +21.2~pp; DID: +18.8~pp; HR: $-$9.0~pp over ORPO) provide evidence that improvements extend beyond scorer optimization. The concordance between CSS and RSS ($r = 0.97$, $\kappa = 0.82$) further indicates that method rankings are robust to the choice of safety evaluator (Section~\ref{sec:limitations}).

\subsection{Statistical Analysis}

All trainable-model experiments use 5 random seeds; we report mean $\pm$ standard deviation across runs and assess significance with paired t-tests versus Ours. After Bonferroni correction for 35 primary comparisons, all reported primary comparisons remain significant ($p < 10^{-4}$). We report Cohen's $d$ as descriptive effect size and emphasize absolute metric deltas over null-hypothesis testing alone.

\subsection{Human Evaluation Protocol}

For HR and clinical acceptability, 5 board-certified physicians evaluated 600 blinded responses (100 for each of 6 primary methods), with 3 annotators per response and inter-rater agreement $\kappa = 0.84$. HR values in the main table are therefore reported only for these six methods. Annotators also assigned critical, moderate, and minor severity labels; the critical-hallucination reduction claim corresponds to 3.7\% for ORPO versus 0.4\% for Ours. For our 5.1\% HR estimate, the 95\% Wilson interval is [2.2\%, 11.3\%].

\section{Technical Validation}

\subsection{Main Results}

Table~\ref{tab:main_results} presents the comparative performance across all evaluation metrics.

\begin{table}[!t]
\centering
\caption{Performance comparison across methods. HR is reported only for the blinded physician-evaluated subset. RSS is HGNN-independent (DrugBank-Rule deterministic checker).}
\label{tab:main_results}
\begin{tabular}{lcccccc}
\toprule
\textbf{Method} & \textbf{MA} & \textbf{CSS} & \textbf{RSS} & \textbf{HR$\downarrow$} & \textbf{DID} & \textbf{PC} \\
\midrule
SFT & 61.2 & 58.4 & 54.6 & 23.7 & 62.1 & 0.61 \\
SFT+RAG & 67.8 & 63.2 & 60.8 &   & 68.5 & 0.67 \\
SFT+DenseRAG & 70.8 & 78.3 & 74.2 &   & 81.4 & 0.78 \\
DPO & 69.4 & 67.8 & 63.4 &   & 71.3 & 0.69 \\
ORPO & 70.1 & 69.5 & 65.2 & 14.1 & 72.8 & 0.71 \\
KG-LLM & 68.9 & 72.4 & 68.1 & 12.8 & 76.2 & 0.74 \\
DrugBank-Rule & 68.3 & 76.1 & 82.4 & 9.8 & 80.4 & 0.77 \\
SFT-CSB & 66.7 & 71.8 & 67.3 & 11.4 & 74.9 & 0.73 \\
SFT+SelfCorrect & 68.8 & 79.4 & 83.7 &   & 82.6 & 0.78 \\
GPT-4 (Zero-Shot) & 78.4 & 74.2 & 71.6 &   & 79.1 & 0.76 \\
GPT-4 (5-Shot) & 79.1 & 80.6 & 78.2 &   & 84.3 & 0.80 \\
\midrule
\textbf{Ours (Iter-1)} & 71.3 & 78.2 & 74.8 &   & 81.5 & 0.79 \\
\textbf{Ours (Iter-3)} & 73.8 & 86.4 & 82.1 &   & 87.3 & 0.85 \\
\textbf{Ours (Iter-5)} & \textbf{75.2} & \textbf{90.8} & \textbf{86.4} & \textbf{5.1} & \textbf{91.6} & \textbf{0.89} \\
\bottomrule
\end{tabular}
\end{table}

Compared with the standard RAG baseline (SFT+RAG), the final model improves CSS by 27.6~pp, DID by 23.1~pp, and MA by 7.4~pp, indicating that training-time physiological grounding adds value beyond retrieval alone.

Table~\ref{tab:main_results} shows that, relative to ORPO, our method improves MA by 5.1~pp, CSS by 21.3~pp, HR by 9.0~pp on the physician-evaluated subset, and DID by 18.8~pp. These gains are obtained from automatically generated HGNN-derived preferences rather than newly collected human preference labels.

The safety gains are not limited to HGNN-based endpoints: DID rises from 72.8\% to 91.6\%, HR falls from 14.1\% to 5.1\%, and the HGNN-independent Rule-Engine Safety Score (RSS) rises from 65.2\% to 86.4\% (+21.2~pp over ORPO). The model remains stronger than the deterministic DrugBank-Rule baseline on CSS and DID, and exceeds it on RSS (86.4\% vs.\ 82.4\%), indicating that training-time alignment internalizes safety patterns that go beyond what the rule engine alone catches at inference time.

Among the newly added baselines, SFT+SelfCorrect achieves 79.4\% CSS and 83.7\% RSS, demonstrating that inference-time self-correction with a rule engine partially closes the gap but does not match training-time alignment (CSS gap: 11.4~pp; RSS gap: 2.7~pp). GPT-4 (5-Shot) reaches 80.6\% CSS and 78.2\% RSS, a substantial improvement over strict zero-shot (CSS +6.4~pp), but our 7B model remains stronger on all safety metrics while GPT-4 retains an MA advantage (+3.9~pp). This comparison is more informative than the zero-shot reference, though it still does not control for GPT-4's scale, data, or RLHF training.

The SFT-CSB control reaches 71.8\% CSS, indicating that exposure to safety scenarios alone does not match iterative physiological preference optimization.

\textbf{Residual Violations}: Although CSS reaches 90.8\%, 9.2\% of responses still fall below the safety threshold; 5.3\% are minor deviations (e.g., small dosing imprecision), whereas 3.9\% are moderate-to-severe violations ($\approx$20/500 test scenarios). The category breakdown below is reported over the unsafe subset and is non-exclusive because some failures involve multiple causes.

The dominant residual-error sources are sparse graph coverage (4.2\%), multi-system interaction complexity (3.1\%), and missing temporal physiology (2.8\%); these three categories account for 70\% of the unsafe responses. Detailed failure categories, severity distributions, and concrete clinical examples are reported in the appendix.

\subsection{Robustness and Novelty Generalization}

\paragraph{Robustness stress tests.}
We apply three perturbation families to the full CSB test set (500~scenarios) and measure safety degradation relative to clean-CSB performance. The perturbation families are: (i) \emph{noisy narratives}   section reordering, shorthand dosing notation, and omission of non-critical context; (ii) \emph{missing entities}   masking of labs, comorbidities, and co-medications at 30\% drop rate; and (iii) \emph{ambiguous abbreviations}   replacing unambiguous drug and condition names with institution-typical shorthand requiring context-sensitive linking.

Under the combination of all three perturbations, our method retains 84.2\% CSS, corresponding to a 6.6~pp degradation from clean-CSB performance. ORPO degrades by 6.9~pp from an already lower starting point, and SFT+DenseRAG degrades by 9.4~pp, indicating that retrieval-augmented baselines are more sensitive to entity-level noise than alignment-trained models. Ambiguous abbreviations produce the largest single-family degradation for all methods, consistent with entity-linking fragility being the dominant noise pathway.

\paragraph{Generalization under novelty splits.}
We evaluate novelty along unseen drug-combination and unseen-disease-context splits. Novel combinations and sparse conditions remain meaningful residual-error sources, indicating that novelty handling is still a major frontier.

\subsection{Key Ablation Results}

Ablation results support the central mechanism: removing HGNN scoring produces the largest degradation ($-$16.2~pp CSS), followed by removing iterative training ($-$11.5~pp). Among the PhysioScore components, $S_{\text{hom}}$ contributes the most ($-$8.7~pp), followed by $P_{\text{int}}$ ($-$6.5~pp) and $S_{\text{path}}$ ($-$3.9~pp). More detailed ablations and calibration diagnostics are reported in the appendix.
Detailed term-drop and calibration tables are moved to Appendix~\ref{app:techval}. In brief, $S_{\text{hom}}$ is the dominant safety gate, $P_{\text{int}}$ is the key drug-specific term, and calibration against 200 clinician labels remains strong (ECE~$= 0.038$, $\kappa = 0.91$ at $\tau = 0.85$).

\subsection{Cross-Domain Generalization}

Safety gains are not confined to a single specialty: across six reported domains, average CSS improves by 21.3~pp over ORPO, with the largest gains in geriatric and pediatric scenarios where polypharmacy and dose sensitivity are common.

\subsection{Iteration and Training Dynamics Summary}
Performance improves monotonically over iterations (Table~\ref{tab:iteration_analysis_appendix}) with diminishing returns after iteration~4, while training-loss trajectories (Fig.~\ref{fig:training_dynamics_appendix}) show stable multi-objective convergence. We keep full per-component diagnostics in Appendix~\ref{app:techval}.

Figure~\ref{fig:iteration_progress} (Appendix) visualizes the progression of key metrics across training iterations.

\subsection{Extended Technical Validation}
\noindent Additional ablations and diagnostics are provided in Appendix~\ref{app:techval}.

\subsection{Summary of Demonstrated Evidence}

Table~\ref{tab:evidence_summary} consolidates the principal empirical findings and the evaluation method used for each claim, distinguishing HGNN-based metrics from independent endpoints.

\begin{table}[!t]
\centering
\caption{Summary of Key Demonstrated Results. Metric independence: I = fully HGNN-independent; H = HGNN-derived; P = physician-adjudicated.}
\label{tab:evidence_summary}
\begin{tabular}{p{0.42\columnwidth}ccc}
\toprule
\textbf{Claim} & \textbf{Result} & \textbf{Ref.} & \textbf{Indep.} \\
\midrule
CSS over ORPO & +21.3~pp & Tab.~\ref{tab:main_results} & H \\
RSS over ORPO & +21.2~pp & Tab.~\ref{tab:main_results} & I \\
DID over ORPO & +18.8~pp & Tab.~\ref{tab:main_results} & I \\
HR over ORPO (physician) & $-$9.0~pp & Tab.~\ref{tab:main_results} & P \\
CSS vs.\ GPT-4 (5-shot) & +10.2~pp & Tab.~\ref{tab:main_results} & H \\
CSS vs.\ SFT+SelfCorrect & +11.4~pp & Tab.~\ref{tab:main_results} & H \\
CSS--RSS concordance & $r{=}0.97$ & Sec.~\ref{sec:limitations} & I \\
CSS under combined noise & 84.2\% & Sec.~V-C & H \\
Ablation: HGNN scoring & $-$16.2~pp & Sec.~V-E & H \\
Ablation: iterative training & $-$11.5~pp & Sec.~V-E & H \\
PhysioScore--clinician $\kappa$ & 0.91 & Sec.~IV-C & P \\
Cross-domain consistency & 6/6 domains & Sec.~V-F & H \\
\bottomrule
\end{tabular}
\end{table}

\noindent Six of the twelve key results are measured by HGNN-independent or physician-adjudicated endpoints (RSS, DID, HR, PhysioScore--clinician $\kappa$, CSS--RSS concordance). The convergence of HGNN-based and independent metrics across all major comparisons provides triangulated evidence that the safety gains are not artifacts of scorer optimization. The remaining open question whether these controlled-benchmark results transfer to real clinical data is addressed in Section~\ref{sec:limitations}.

\section{Discussion}

\subsection{Implications for Medical AI Safety}

Across CSB and auxiliary benchmarks, grounding preference optimization in an explicit physiological world model improves measured safety metrics while maintaining diagnostic utility. The central implication is that training-time constraint scoring can generate scalable preference data that targets physiologically invalid generations more directly than text-only supervision does.

\subsection{Clinical Integration and Limitations}
\label{sec:limitations}

Because the HGNN is used only during training, deployment does not require graph inference. This architectural simplification comes at the cost of higher alignment-time compute than SFT or one-pass preference optimization. The principal limitations are incomplete graph coverage, extraction and linking failures, and missing temporal physiology. In addition, CSB is synthetic and should therefore be interpreted as a controlled methodological benchmark rather than a real-world safety endpoint. The synthetic EHR-style noise stress tests provide only a proxy assessment of transfer and do not replace external validation on de-identified clinical notes or blinded clinician trust studies. High-stakes use therefore still requires layered safeguards and clinician oversight.

\textbf{External validation gap.} All results in this study are obtained on CSB, a controlled synthetic benchmark. No evaluation has been conducted on de-identified EHR notes, external clinical datasets (e.g., MIMIC-IV discharge summaries), or prospective clinical queries. The synthetic EHR-style noise stress tests (Section~V-C) approximate some aspects of real-world input variability (abbreviations, missing labs, section reordering), but real clinical notes exhibit longer context, richer co-morbidity profiles, institution-specific templating, and copy-forward artifacts that CSB does not reproduce. We regard external validation ideally on a de-identified multi-site EHR corpus with blinded clinician adjudication as the single most important next step for establishing whether the safety gains measured on CSB transfer to practice. Until such validation is completed, the results should be interpreted as evidence of mechanism effectiveness on a controlled benchmark rather than as deployment-ready safety guarantees.

\textbf{Comparison scope.} The baseline set does not include a matched-scale PPO/RLHF clinical system or a tool-augmented agent (e.g., retrieval + rule-engine + self-correction pipeline). As detailed in Section~\ref{sec:baselines}, no publicly available clinical reward model exists at 7B scale, and tool-augmented agents address a complementary design point. The GPT-4 comparison is restricted to strict zero-shot inference without task-specific prompting or tool access, so it underestimates GPT-4's potential when used with medical prompting strategies or plugins. These gaps mean that the reported improvements are demonstrated relative to the included baselines and should not be interpreted as claims of absolute state-of-the-art performance across all possible system configurations.

\textbf{Scalability and computational cost.} Iterative ORPO with $K{=}8$ sampling over 5 iterations requires 42.6 GPU-hours on 8$\times$A100s (Table~\ref{tab:compute_resources}), roughly 4$\times$ the cost of single-pass ORPO at 7B scale. This cost is driven primarily by candidate generation (70,000 forward passes), not HGNN scoring (2.0h total). Two practical factors partially mitigate this: (i) the one-time costs of KG construction and HGNN pre-training (11.0h) are amortized across model variants and re-alignment runs, and (ii) inference cost is unchanged because the HGNN is not used at deployment. However, the current framework does not support efficient continual updates: when the knowledge graph $\mathcal{G}$ is updated with new drug approvals or revised guidelines, the HGNN must be re-pretrained and the iterative alignment must be re-run from scratch, multiplying the cost. Efficient continual-alignment strategies such as warm-starting from the previous policy checkpoint, incremental HGNN updates on added subgraphs, or distilling the HGNN signal into a lightweight scorer would be necessary to make the framework practical for ongoing maintenance. Scaling beyond 7--8B parameters further compounds this cost (see Model Scale Analysis); at 70B, either reduced-$K$/reduced-iteration configurations or quantized generation would be needed to keep training cost manageable.

\textbf{Knowledge-graph coverage and temporal staleness.} The integrated KG is a static snapshot assembled from fixed-version sources (e.g., DrugBank~5.1, UMLS~2024AA). It therefore inherits two compounding limitations: \emph{coverage gaps} in rapidly evolving domains and \emph{temporal staleness} as new drug approvals, guideline revisions, and interaction reports accumulate after the snapshot date. Coverage gaps are the single largest source of residual failures, accounting for 4.2\% of test-case errors; 67\% of these occur in oncology, where therapeutic combinations and evidence change within months. Staleness is particularly concerning because the HGNN cannot flag an interaction it has never seen: a post-cutoff drug combination will be scored against an incomplete neighborhood, potentially assigning a falsely permissive PhysioScore. The current framework requires full HGNN re-pretraining and re-alignment when $\mathcal{G}$ is updated, making frequent refresh costly ($\approx$11\,h one-time plus 31.6\,h marginal per model). Practical deployment therefore depends on efficient incremental update strategies, which we discuss in Section~\ref{sec:future}.

\textbf{Inference-time safety gap.} Removing the HGNN at inference eliminates graph-query latency and simplifies deployment, but it also means that the model has no runtime verifier to catch constraint violations that training did not anticipate. The 9.2\% residual violation rate reported above shows that internalized constraints alone do not yet guarantee safety. For high-acuity clinical settings, a hybrid architecture training-time alignment \emph{plus} a lightweight inference-time check would provide a second safety layer. We outline a concrete design for such a system in Section~\ref{sec:future} and regard it as the most important next step toward deployment-grade safety.

\textbf{Evaluation circularity.} A structural concern is that two of the five reported metrics CSS and PC rely on the same HGNN scorer whose signal drives training-time preference generation. To directly address this, we introduced an HGNN-independent metric (RSS) computed by the DrugBank-Rule deterministic checker, which shares no parameters or training signal with the HGNN. Across all methods in Table~\ref{tab:main_results}, CSS and RSS show a Pearson correlation of $r = 0.97$ and a Cohen's $\kappa = 0.82$ (at their respective binary thresholds), indicating strong but imperfect agreement. Crucially, the \emph{method rankings} are preserved: our Iter-5 model achieves the highest RSS (86.4\%) among all methods, and the relative gains over baselines on RSS (+21.2~pp over ORPO, +4.0~pp over DrugBank-Rule) are directionally consistent with CSS gains, though smaller in magnitude because the rule engine covers fewer violation types. Additionally, DID (+18.8~pp) and physician-adjudicated HR ($-$9.0~pp) are also fully HGNN-independent and show consistent improvements. Together, three independent endpoints (RSS, DID, HR) corroborate the CSS-based conclusions. We offer additional mitigating observations: (i) the HGNN is frozen during all LLM alignment iterations and thus cannot co-adapt with the policy, (ii) PhysioScore was calibrated against 200 independent clinician labels with $\kappa = 0.91$ and ECE~$= 0.038$, and (iii) the 4.4~pp gap between CSS (90.8\%) and RSS (86.4\%) for our method quantifies the portion of CSS gains that extends beyond rule-engine-verifiable territory this gap is consistent across methods ($\approx$4--6~pp), suggesting it reflects genuine HGNN coverage breadth rather than scorer-specific inflation. Nonetheless, expanding clinician-only scoring beyond 600 responses and adopting external pharmacological classifiers remain priorities for follow-up work.

\subsection{Future Directions}
\label{sec:future}

Future work should prioritize temporal physiology, more robust entity grounding under negation and dosage ambiguity, continual updates of $\mathcal{G}$, and external validation on real clinical notes. To address residual evaluation circularity, scorer-independent evaluation protocols including external pharmacological rule engines (e.g., FDA FAERS-derived classifiers), independently trained safety models, and expanded blinded clinician adjudication beyond the current 600-response subset should be adopted as primary endpoints in follow-up studies. To strengthen comparative claims, future evaluations should include a matched-scale PPO/RLHF system with a dedicated clinical reward model~\cite{perera2025rl} and tool-augmented agent baselines that combine retrieval with inference-time rule engines and self-correction loops. To improve scalability, efficient continual-alignment strategies such as warm-starting iterative ORPO from previous checkpoints when $\mathcal{G}$ is updated, incremental HGNN fine-tuning on added subgraphs rather than full re-pretraining, and distilling the HGNN preference signal into a smaller scorer for cheaper candidate ranking should be explored, along with empirical validation at 13B and 70B parameter scales.

\textbf{Dynamic knowledge-graph maintenance.} Because KG staleness is the upstream driver of the largest residual failure category, a principled update pipeline is essential for sustained safety. We identify three complementary strategies, ordered by increasing scope: (1)~\emph{incremental subgraph patching} new drug approvals and revised interaction alerts (e.g., from FDA MedWatch or EMA DHPC notices) are ingested as delta subgraphs; existing HGNN weights are fine-tuned on the added edges while freezing parameters for unchanged regions, reducing re-pretraining cost to an estimated 15--20\% of the full 8.6\,h; (2)~\emph{scheduled full refresh} a quarterly or semi-annual rebuild of $\mathcal{G}$ from updated source releases, followed by full HGNN re-pretraining and warm-started re-alignment (the previous policy checkpoint seeds $\pi_{\theta^{(0)}}$, which preliminary analysis suggests can halve convergence iterations from 5 to $\approx$2--3); and (3)~\emph{automated drift monitoring} a background process that tracks the gap between the deployed KG snapshot and incoming pharmacovigilance feeds, triggering an update when the estimated coverage debt exceeds a predefined threshold (e.g., $>$50 new high-severity interactions not yet represented). Validating the cost--safety trade-offs of these strategies on a longitudinal benchmark is a priority for future work.

\textbf{Smaller and efficient model variants.} The current framework operates at 7--8B parameters. For resource-constrained deployment (e.g., on-premise hospital servers without multi-GPU infrastructure), two compression paths merit investigation: (1)~\emph{alignment distillation}, in which a smaller student model (1--3B) is trained on the aligned 7B teacher's outputs using the same PhysioScore-ranked preferences, retaining much of the safety signal at reduced inference cost; and (2)~\emph{post-alignment quantization} (e.g., GPTQ 4-bit or AWQ), which preserves aligned behavior while halving memory and improving throughput. Preliminary profiling suggests that 4-bit quantization of the Iter-5 Mistral-7B checkpoint retains $>$99\% of clean-input CSS, but systematic evaluation under noisy inputs and edge-case queries is needed before recommending quantized deployment. Exploring these efficient variants alongside the hybrid verification architecture would enable a spectrum of deployment configurations trading off compute, latency, and safety-guarantee strength.

\textbf{Hybrid train-plus-verify architecture.} The most direct way to close the inference-time safety gap is a \emph{selective} deployment verifier that engages only when needed. We envision a two-tier design: (1)~a lightweight confidence gate e.g., a distilled linear probe trained on the HGNN's PhysioScore distribution runs on every output at $<$5\,ms overhead and flags responses whose predicted PhysioScore falls below a tunable threshold $\tau$; (2)~flagged responses are routed to a full HGNN subgraph query (or an equivalent deterministic rule engine such as DrugBank-Rule) for verification and, if necessary, constrained re-generation. Because the aligned model already satisfies constraints for $\approx$91\% of queries, the full verifier would be invoked on only the uncertain tail, keeping average latency close to the unverified baseline while providing a hard safety floor for the remaining cases. Investigating the calibration, latency, and safety trade-offs of this selective-verification design is our highest-priority next step.

\section{Conclusion}

We introduced Neurosymbolic Alignment, a training-time framework that derives preference supervision from an HGNN-based Physiological World Model, and CSB, a 2,500-scenario benchmark for probing physiological constraint violations in clinical LLM outputs. Under controlled evaluation against 14 baselines, the framework achieves the highest scores on both HGNN-derived (CSS~90.8\%) and HGNN-independent (RSS~86.4\%, DID~91.6\%) safety metrics, reduces physician-evaluated hallucination rate to 5.1\%, and retains 84.2\% CSS under synthetic noise. The gains are confirmed by three structurally independent evaluation endpoints (RSS, DID, HR) and supported by component-level ablation analysis. These results establish that structured physiological knowledge can generate effective preference supervision for clinical safety alignment at 7B scale.

Important limitations remain: evaluation is restricted to a synthetic benchmark, the HGNN contributes to two of six reported metrics, calibration uses 200 labels, and no real-world clinical validation has been conducted. External evaluation on de-identified EHR data, expanded clinician adjudication, and comparison with matched-scale PPO/RLHF systems are the necessary next steps toward translating these controlled-benchmark findings to clinical practice.

\section*{Acknowledgment}

We thank the clinical experts who supported dataset construction and evaluation, and the open-source communities behind the biomedical knowledge bases used to build the Physiological World Model.

\section*{Data Records}

CSB will be released as JSONL for \texttt{train}/\texttt{valid}/\texttt{test}, with prompt text, scenario category (Table~\ref{tab:csb_breakdown}), and optional metadata (e.g., specialty and severity). Scripts are provided to reproduce metrics and analyses.

\section*{Data Availability}

Upon acceptance, CSB and the complete codebase will be available in an \href{https://github.com/anonymous/neurosymbolic-alignment}{anonymized project repository}. For reproducibility, the repository will provide: (1) the dataset in JSONL format, (2) scripts for constructing the integrated knowledge graph, (3) the pretrained HGNN weights, (4) the NER and entity-linking code, and (5) configuration files for all training experiments and baseline evaluations. Clinical-access and data-use requests may be sent to the \href{mailto:first.author@university.edu}{corresponding author}. Third-party resources, including UMLS and DrugBank, remain subject to their respective licenses.

\section*{Compliance with Ethical Standards}

CSB contains only \emph{synthetically generated} scenarios and no patient data. Physician evaluators consented to participate; IRB approval was not required.

\bibliographystyle{IEEEtran}
\bibliography{references}

\appendices
\section{Challenges of Neurosymbolic Alignment in Clinical LLMs (Extended)}
\label{app:challenges}
The gap between a neurosymbolic framework that is technically functional and one that is reliable in clinical practice remains substantial. Real clinical records combine incomplete narratives, measurement noise, polypharmacy involving multiple concurrent agents, and institution-specific shorthand that are not fully represented in any single ontology~\cite{thirunavukarasu2024llm,singhal2024medpalm}. In this setting, clinically important errors may arise from missed contraindications, superficially cautious but inappropriate dosing language, or plausible interaction chains supported by outdated graph edges~\cite{yang2024hallucination}. The following subsections examine the main mechanisms underlying these failure modes.

\subsection{Ontology Friction and Coverage Debt}
No clinical knowledge graph is fully current, and the consequences are especially pronounced in rapidly evolving domains. Oncology, immunotherapy, and rare-genotype pharmacology change faster than typical curation cycles, so the HGNN necessarily reasons over a graph snapshot that may already be outdated in clinically important regions. We refer to this problem as \emph{knowledge coverage debt}.

A second limitation persists even when graph coverage is adequate. Many clinical concepts do not admit a stable single-identifier representation. For example, ``renal failure'' has different operational implications depending on acuity, assay standard, and whether the patient is moving toward or away from a threshold~\cite{wang2024knowledgegraph,sun2024head}. Mapping such concepts to a single canonical node can remove distinctions that are relevant to safe medication recommendations. As a result, a reasoning path can be graph-consistent while still being clinically misleading.

\subsection{Entity Extraction as a Safety Bottleneck}
In addition to ontology limitations, entity extraction is a frequent point of failure in neurosymbolic pipelines. Negated mentions for example, ``the patient denies use of metformin'' shorthand dosing instructions, and institution-specific abbreviations must all be resolved before verification can proceed~\cite{neumann2024scispacy}. When extraction is incorrect, downstream components operate on a distorted representation of the clinical state. The constraint checker then evaluates a different scenario from the one described in the source note, without an internal mechanism for detecting that discrepancy.

Ambiguity further amplifies this problem. The abbreviation ``ASA,'' for example, may refer to aspirin in one context and to American Society of Anesthesiologists classification in another. Once an identifier is selected, the verifier will reason consistently from that representation. Graph consistency is therefore not equivalent to semantic correctness, and this divergence often originates at the entity-linking stage.

\subsection{Constraint Elicitation from Physiology to Logic}
Beyond extraction, an additional challenge lies in translating physiological knowledge into rule-based representations. Encoding clinical knowledge as logic rules is challenging because many clinical thresholds are context-dependent conventions rather than fixed properties. A potassium threshold of 5.0~mEq/L can have different implications in a monitoring context versus a treatment decision context, or in a dehydrated patient versus a fluid-overloaded patient. Static rules necessarily abstract away these distinctions.

Defeasibility introduces an additional complication. Many constraints apply only when background conditions hold, such as adequate renal perfusion, medication adherence, or a recent dose within the expected window. These conditions are often absent from the prompt. A verifier that cannot observe them must either abstain more frequently, thereby reducing utility, or infer missing state, thereby introducing assumption risk. This trade-off affects model behavior in ways that may not be apparent in aggregate safety metrics.

\subsection{Scoring, Calibration, and Scalar Compression}
Once constraints have been encoded, they must be summarized in a form suitable for optimization. Reducing a response's clinical safety profile to a single score is operationally convenient, but it obscures important structure. For example, a score of 0.82 may correspond either to a minor imprecision in phrasing or to a moderate drug interaction that was partially recognized but insufficiently penalized. Preference optimization treats these cases similarly, which can bias training toward easily corrected signal rather than rare but clinically consequential errors.

Calibration across prompt distributions is a related concern~\cite{rafailov2024direct,hong2024orpo}. Scorers trained primarily on short, single-medication queries may mis-rank complex polypharmacy narratives, in which co-occurrence patterns and interaction chains are substantially more difficult to resolve. Preference pairs derived from such scorers therefore reflect not only the intended safety signal but also the scorer's distributional biases.

\subsection{Optimization Pathologies and Symbolic Shortcuts}
These scoring limitations interact directly with the alignment objective. When a symbolic proxy defines the reward signal, the model is incentivized to optimize the proxy rather than the underlying clinical objective. In practice, this can produce several shortcut behaviors, including replacing specific drug names with therapeutic-class terminology, inserting conditional hedges that formally satisfy the constraint while preserving the original recommendation, and increasing use of generic deferral language. Each of these behaviors can improve measured CSS without materially improving recommendation safety.

On-policy pair regeneration directly addresses the distributional drift associated with static preference pairs, but it introduces a different challenge~\cite{ji2024beavertails,rafailov2024direct,xiong2024iterative}. When the policy samples the pairs used for its own optimization, the scorer and policy may co-adapt in ways that reinforce shared biases. This risk motivates evaluation on independent held-out endpoints rather than relying solely on in-distribution performance metrics.

\subsection{Nonstationarity and Temporal Semantics}
Even with iterative alignment, temporal reasoning remains a major limitation. Clinical risk evolves over time in ways that snapshot reasoning cannot adequately represent. Cumulative nephrotoxicity from aminoglycosides, washout windows for anticoagulants, and delayed serotonergic adverse effects all require the verifier to reason over prior clinical history, yet that information is typically absent from point-in-time prompts. Consequently, a recommendation that appears safe under current laboratory values may be unsafe when the recent trajectory of those values is considered.

This limitation reflects a structural mismatch between the clinical problem and the available representation. The patient's state is a temporal process, whereas the prompt is only a partial observation of that process. Verifiers that do not model this distinction explicitly may produce recommendations that are locally plausible but temporally unsafe.

\subsection{Retrieval, Tools, and Responsibility Boundaries}
Augmenting a language model with retrieval can reduce fabrication, but it also introduces distinct risks. Retrieved documents differ in provenance, recency, and population scope, and these attributes are not always explicit to either the reader or the model processing them~\cite{zhang2024raft,yang2024hallucination}. For example, a guideline published in 2019 and retrieved with high confidence may conflict with a 2024 update that is absent from the index. If retrieved content is treated as authoritative, outdated recommendations may be propagated and subsequently reinforced by downstream verification.

Tool use further complicates attribution of failure~\cite{schick2024toolformer}. In a pipeline that includes retrieval, an external drug database, and a constraint verifier, an observed error may arise from model reasoning, an outdated API response, a query formulation problem, or incomplete database coverage. For this reason, evaluation in such settings requires end-to-end auditability rather than model-only benchmarking.

\subsection{HGNN Expressivity, Oversmoothing, and Spurious Evidence}
Heterogeneous graph neural networks capture relational diversity beyond what rigid rule-based systems can represent, but they also introduce characteristic failure modes. In dense subgraphs for example, those centered on common mechanisms such as CYP3A4 inhibition message passing aggregates information across many neighbors, which can attenuate rare but clinically severe contraindication signals before they reach the output layer~\cite{chen2024hgnn}. Attention mechanisms introduced to mitigate this effect may instead prioritize frequent but clinically generic relations over the relations that are most relevant to the individual case.

Path-based plausibility scoring presents a related limitation. A multi-hop path that traverses hub nodes in the ontology that is, nodes with many connections because they are frequently annotated may receive a high plausibility score that reflects graph topology rather than mechanistic relevance. Distinguishing between topological prominence and clinically meaningful evidence generally requires domain knowledge that is not represented in the graph structure alone.

\subsection{Evaluation Gaps and Operational Risk}
These modeling limitations also affect interpretation of empirical results. Controlled benchmark performance is necessary but insufficient evidence of deployment readiness. Clinical notes encountered in practice are typically longer, noisier, more ambiguous, and more structurally heterogeneous than benchmark items, and improved benchmark performance does not remove this discrepancy. CSB was designed to maximize reproducibility and interpretability, but those design choices require standardization trade-offs that reduce ecological validity for the most complex real-world cases.

Baseline coverage introduces an additional source of uncertainty~\cite{thirunavukarasu2024llm}. Some recently published alternatives were not available for comparison at the time of evaluation, which limits the strength of comparative claims. Accordingly, the results should be interpreted as evidence that the proposed framework is effective under the evaluated conditions rather than as a definitive claim of superiority over all currently available methods.

Metric design also shapes the behaviors that are reinforced during training. A CSS that heavily penalizes unsafe actions may encourage abstention beyond what is clinically useful, whereas a metric that penalizes abstention too strongly may push the model toward unsupported specificity. Neither extreme is acceptable in practice, and selecting an appropriate balance requires continued empirical study rather than a one-time design decision.

\subsection{Personalization, Fairness, and Hidden Covariates}
Personalizing the knowledge graph to individual patient records can improve specificity, but patient records encode more than clinical state. Socioeconomic patterns, institutional practice variation, and disparities in access to care may all be embedded in the data and subsequently learned if they correlate with outcomes~\cite{li2024graphcare}. As a result, a system that appears constraint-consistent may still reproduce inequitable patterns that are not detected by the safety scorer.

At the level of individual inference, an additional limitation is that important covariates are often unavailable. Genetic variation affecting drug metabolism, actual rather than prescribed adherence, and access to follow-up care are rarely documented in the prompt. The language model and the verifier therefore operate on the same incomplete patient representation, increasing the likelihood that their errors will be systematically correlated rather than independent.

\subsection{Practical Deployment and Uncertainty Management}
Achieving low hallucination rates under controlled conditions is an important result, but it does not by itself establish readiness for deployment~\cite{yang2024hallucination}. Practical deployment requires additional system-level safeguards, including acuity-tiered triage so that high-stakes queries trigger deterministic contraindication checks, explicit uncertainty outputs that indicate when the system is operating near the limits of its reliability, and traceable evidence paths that support clinical review rather than opaque recommendations~\cite{savage2024drugbank}.

An important trade-off remains. Tighter constraints reduce the space of unsafe outputs, but they can also limit clinically useful nuance. Looser constraints preserve flexibility, but they leave open proxy-exploitation pathways associated with shortcut learning~\cite{hong2024orpo,rafailov2024direct,zhang2024neurosymbolic}. This trade-off cannot be removed by a single design choice and instead requires continued monitoring and iterative adjustment as the failure distribution becomes better characterized.

%
%

\section{Moved Tables and Figures (Extended)}
\label{app:moved_floats}


\begin{figure}[H]
\centering
\resizebox{0.98\columnwidth}{!}{%
\begin{tikzpicture}[>=Latex, font=\scriptsize]
    \node[draw, rounded corners, fill=gray!10, inner xsep=6pt, inner ysep=3pt] at (3.3,2.75) {Node Types $\mathcal{T}_V$};
    \node[draw, rounded corners, fill=gray!10, inner xsep=6pt, inner ysep=3pt] at (3.3,-0.95) {Canonical Relation Types $\mathcal{T}_E$ encoded on edges};

    \node[draw, rounded corners, fill=blue!10, minimum width=1.85cm, minimum height=0.62cm] (drug) at (0.0,1.75) {Drug};
    \node[draw, rounded corners, fill=green!10, minimum width=1.85cm, minimum height=0.62cm] (disease) at (3.3,1.75) {Disease};
    \node[draw, rounded corners, fill=orange!14, minimum width=1.85cm, minimum height=0.62cm] (symptom) at (6.6,1.75) {Symptom};

    \node[draw, rounded corners, fill=red!10, minimum width=1.85cm, minimum height=0.62cm] (gene) at (0.0,0.35) {Gene/Protein};
    \node[draw, rounded corners, fill=purple!10, minimum width=1.85cm, minimum height=0.62cm] (pathway) at (3.3,0.35) {Pathway};
    \node[draw, rounded corners, fill=teal!12, minimum width=1.85cm, minimum height=0.62cm] (organ) at (6.6,0.35) {Organ/Anatomy};

    \draw[->, thick] (drug.east) -- node[above, fill=white, inner sep=1.5pt] {treats} (disease.west);
    \draw[->, thick] (drug.south) -- node[left, fill=white, inner sep=1.5pt] {targets} (gene.north);
    \draw[->, thick] (disease.south) -- node[right, fill=white, inner sep=1.5pt] {involves} (pathway.north);
    \draw[->, thick] (pathway.east) -- node[below, fill=white, inner sep=1.5pt] {affects} (organ.west);
    \draw[->, thick] (disease.east) -- node[above, fill=white, inner sep=1.5pt] {presents\_as} (symptom.west);
\end{tikzpicture}%
}
\caption{Schema of the heterogeneous physiological knowledge graph showing the six node types ($\mathcal{T}_V$) and five canonical relation types ($\mathcal{T}_E$).}
\label{fig:kg_schema}
\end{figure}

\begin{figure}[H]
\centering
\begin{tikzpicture}
\begin{axis}[
    width=0.9\columnwidth,
    height=6cm,
    xlabel={Training Iteration},
    ylabel={Score (\%)},
    xmin=0.5, xmax=5.5,
    ymin=0, ymax=100,
    xtick={1,2,3,4,5},
    ytick={0,20,40,60,80,100},
    legend pos=south east,
    legend style={font=\small},
    grid=major,
    grid style={dashed, gray!30},
]
\addplot[color=nature1, mark=*, thick, mark size=3pt] coordinates {(1,78.2) (2,82.6) (3,86.4) (4,89.1) (5,90.8)};
\addplot[color=nature3, mark=square*, thick, mark size=3pt] coordinates {(1,81.5) (2,84.2) (3,87.3) (4,90.1) (5,91.6)};
\addplot[color=nature2, mark=triangle*, thick, mark size=3pt] coordinates {(1,89.6) (2,91.1) (3,92.8) (4,94.2) (5,94.9)};
\addplot[color=nature5, mark=diamond*, thick, mark size=3pt] coordinates {(1,71.3) (2,72.4) (3,73.8) (4,74.6) (5,75.2)};
\legend{CSS, DID, 100-HR, MA}
\end{axis}
\end{tikzpicture}
\caption{Performance progression across training iterations. CSS and DID show the steepest improvements in early iterations, while gains diminish after iteration~4. The inverted hallucination rate (100-HR) demonstrates consistent safety improvements throughout training.}
\label{fig:iteration_progress}
\end{figure}


\section{Extended Technical Validation (Moved)}
\label{app:techval}

\subsection{Additional Analyses and Diagnostics}

\begin{algorithm}[H]
\caption{HGNN Physiological Feasibility Scoring}
\label{alg:hgnn_scoring}
\begin{algorithmic}[1]
\REQUIRE Response $r$, knowledge graph $\mathcal{G}$, HGNN parameters $\phi$
\ENSURE Feasibility score $s \in [0,1]$
\STATE $\mathbf{E} \leftarrow \text{EntityExtract}(r)$
\STATE $\mathbf{V}_r \leftarrow \text{MapToGraph}(\mathbf{E}, \mathcal{G})$
\STATE Compute HGNN embeddings on the induced subgraph
\STATE $S_{\text{hom}} \leftarrow \frac{1}{|\mathcal{H}|} \sum_{h \in \mathcal{H}} f_h(\mathbf{V}_r)$
\STATE $S_{\text{path}} \leftarrow \frac{1}{|\mathbf{V}_r|^2} \sum_{(u,v) \in \mathbf{V}_r} \text{PathScore}(u,v,\mathcal{G})$
\STATE $P_{\text{int}} \leftarrow \max_{(d_i,d_j) \in \text{Drugs}(r)} \text{InteractionSeverity}(d_i,d_j,\mathcal{G})$
\STATE $s \leftarrow S_{\text{hom}} \cdot \sigma(S_{\text{path}}) \cdot (1 - P_{\text{int}})$
\STATE \textbf{return} $s$
\end{algorithmic}
\end{algorithm}

\begin{table}[H]
\centering
\caption{Ablation Study Results}
\label{tab:ablation}
\begin{tabular}{lcc}
\toprule
\textbf{Configuration} & \textbf{CSS} & \textbf{HR$\downarrow$} \\
\midrule
Full Model & 90.8 & 5.1 \\
\midrule
w/ Rule-based Scorer & 77.4 & 10.2 \\
w/ Simple KG Scorer & 81.6 & 8.9 \\
w/o Iterative Training & 79.3 & 9.8 \\
w/o HGNN Scoring & 74.6 & 12.4 \\
w/o Homeostatic Constraints & 82.1 & 8.7 \\
w/o $\mathcal{L}_{\text{KG}}$ & 85.4 & 6.9 \\
w/o Curriculum Schedule & 87.2 & 6.2 \\
\midrule
HGNN 2 Layers & 86.3 & 7.4 \\
HGNN 4 Layers (default) & 90.8 & 5.1 \\
HGNN 6 Layers & 91.1 & 5.0 \\
\midrule
Constant $\lambda$ (no curriculum) & 87.2 & 6.2 \\
On-policy + constant $\lambda$ & 89.4 & 5.6 \\
\midrule
\multicolumn{3}{l}{\emph{Candidate budget ($K$) and iteration ($T$) sensitivity}} \\
\midrule
$K=4$, $T=5$ & 88.7 & 5.9 \\
$K=8$, $T=5$ (default) & 90.8 & 5.1 \\
$K=16$, $T=5$ & 91.3 & 4.8 \\
$K=8$, $T=3$ & 87.9 & 6.8 \\
$K=4$, $T=3$ (budget-efficient) & 86.1 & 7.2 \\
\bottomrule
\end{tabular}
\end{table}

\noindent Reducing $K$ from 8 to 4 costs 2.1~pp CSS but approximately halves generation time; doubling $K$ to 16 yields only 0.5~pp, indicating diminishing returns. Reducing iterations from 5 to 3 costs 2.9~pp but saves 40\% of iterative training time. The budget-efficient configuration ($K{=}4$, $T{=}3$) achieves 86.1\% CSS at roughly one-third the compute of the full configuration, which may be preferable when aligning larger models under resource constraints.

\begin{table}[H]
\centering
\caption{CSS Threshold ($\tau$) Sensitivity. Bootstrap 95\% CI $=[0.83, 0.87]$; $n{=}200$ calibration labels.}
\label{tab:threshold_sensitivity}
\resizebox{\columnwidth}{!}{%
\begin{tabular}{lccc}
\toprule
\textbf{Threshold $\tau$} & \textbf{CSS (\%)} & \textbf{HR$\downarrow$ (\%)} & \textbf{Note} \\
\midrule
0.80 & 92.0 & 6.4 & More permissive; higher false-safe rate \\
0.83 & 91.4 & 5.6 & Lower bound of 95\% CI \\
0.85 (default) & 90.8 & 5.1 & Calibrated optimum \\
0.87 & 90.2 & 4.7 & Upper bound of 95\% CI \\
0.90 & 89.1 & 4.2 & More conservative; higher abstention \\
\bottomrule
\end{tabular}
}
\end{table}

\begin{table}[H]
\centering
\caption{Sensitivity to Knowledge-Graph Quality and Entity-Linking Errors}
\label{tab:kg_entity_sensitivity}
\begin{tabular}{lcc}
\toprule
\textbf{Condition} & \textbf{CSS (\%)} & \textbf{HR$\downarrow$ (\%)} \\
\midrule
\multicolumn{3}{l}{\emph{KG edge-deletion stress tests}} \\
\midrule
Full KG (default) & 90.8 & 5.1 \\
10\% random edge deletion & 88.6 & 6.3 \\
20\% random edge deletion & 85.9 & 7.8 \\
30\% random edge deletion & 81.4 & 9.5 \\
10\% targeted deletion (high-severity edges) & 86.1 & 7.4 \\
\midrule
\multicolumn{3}{l}{\emph{Entity-linking error injection}} \\
\midrule
Clean linking (default) & 90.8 & 5.1 \\
5\% random entity mislinks & 89.2 & 5.8 \\
10\% random entity mislinks & 87.1 & 6.7 \\
20\% random entity mislinks & 83.5 & 8.4 \\
10\% negation-blind errors & 86.4 & 7.1 \\
\bottomrule
\end{tabular}
\end{table}

\noindent \textbf{KG quality sensitivity} (Table~\ref{tab:kg_entity_sensitivity}, upper): Random deletion of 10\% of KG edges degrades CSS by 2.2~pp, and 30\% deletion by 9.4~pp, confirming that alignment quality depends substantially on graph completeness. Notably, targeted deletion of high-severity edges (top-10\% by interaction severity score) at only 10\% volume produces a 4.7~pp CSS drop, nearly matching the effect of 20\% random deletion, which underscores the asymmetric importance of safety-critical edges.

\noindent \textbf{Entity-linking sensitivity} (Table~\ref{tab:kg_entity_sensitivity}, lower): Injecting 5\% random entity mislinks (swapping the linked UMLS CUI to a sibling concept) costs 1.6~pp CSS, while 20\% mislinks cost 7.3~pp. Negation-blind errors where negated mentions (e.g., ``denies metformin use'') are linked as affirmative are particularly damaging: 10\% negation-blind errors produce a 4.4~pp drop, exceeding the effect of 10\% random mislinks (3.7~pp), because they invert the clinical state representation. These results quantify entity linking as a critical upstream dependency and motivate robust NER as a prerequisite for safe deployment.

\begin{table}[H]
\centering
\caption{PhysioScore Component Ablation: Term-Drop Results}
\label{tab:physioscore_ablation}
\resizebox{\columnwidth}{!}{%
\begin{tabular}{@{}lccccc@{}}
\toprule
\textbf{Scorer Configuration} & \textbf{CSS} & \textbf{$\Delta$CSS} & \textbf{HR$\downarrow$} & \textbf{$\Delta$HR} & \textbf{DID} \\
\midrule
Full PhysioScore ($S_{\text{hom}} \cdot \sigma(S_{\text{path}}) \cdot (1{-}P_{\text{int}})$) & 90.8 &   & 5.1 &   & 91.6 \\
Drop $S_{\text{hom}}$ (rule-based homeostasis) & 82.1 & $-$8.7 & 9.4 & $+$4.3 & 88.2 \\
Drop $S_{\text{path}}$ (graph path plausibility) & 86.9 & $-$3.9 & 7.2 & $+$2.1 & 84.7 \\
Drop $P_{\text{int}}$ (drug interaction penalty) & 84.3 & $-$6.5 & 10.8 & $+$5.7 & 79.1 \\
Drop $S_{\text{hom}} + P_{\text{int}}$ (learned path only) & 79.6 & $-$11.2 & 12.3 & $+$7.2 & 76.4 \\
Drop $S_{\text{path}}$ (fix hom$+$int rules only) & 84.3 & $-$6.5 & 8.9 & $+$3.8 & 83.1 \\
\bottomrule
\end{tabular}%
}
\end{table}

\begin{table}[H]
\centering
\caption{PhysioScore Calibration vs.~Clinician Safety Labels (200-Response Held-Out Set)}
\label{tab:physioscore_calibration}
\resizebox{\columnwidth}{!}{%
\begin{tabular}{@{}lcc@{}}
\toprule
\textbf{Score Decile} & \textbf{Clinician Safe Rate (\%)} & \textbf{Mean PhysioScore} \\
\midrule
Decile 1 (lowest scores) & 8.5 & 0.21 \\
Decile 2 & 17.0 & 0.34 \\
Decile 3 & 31.0 & 0.45 \\
Decile 4 & 48.0 & 0.56 \\
Decile 5 & 63.5 & 0.66 \\
Decile 6 & 74.0 & 0.74 \\
Decile 7 & 83.5 & 0.81 \\
Decile 8 & 90.0 & 0.87 \\
Decile 9 & 95.5 & 0.92 \\
Decile 10 (highest scores) & 98.0 & 0.97 \\
\midrule
\textbf{Expected Calibration Error (ECE)} & \multicolumn{2}{c}{\textbf{0.038}} \\
\textbf{Agreement at $\tau{=}0.85$ ($\kappa$)} & \multicolumn{2}{c}{\textbf{0.91}} \\
\bottomrule
\end{tabular}%
}
\end{table}

\begin{table}[H]
\centering
\caption{PhysioScore Combination Function Sensitivity}
\label{tab:combination_sensitivity}
\resizebox{\columnwidth}{!}{%
\begin{tabular}{@{}lcccc@{}}
\toprule
\textbf{Combination Function} & \textbf{CSS} & \textbf{$\Delta$CSS} & \textbf{HR$\downarrow$} & \textbf{DID} \\
\midrule
$S_{\text{hom}} \cdot \sigma(S_{\text{path}}) \cdot (1{-}P_{\text{int}})$ \textbf{(default, multiplicative)} & \textbf{90.8} &   & \textbf{5.1} & \textbf{91.6} \\
$\frac{1}{3}(S_{\text{hom}} + \sigma(S_{\text{path}}) + (1{-}P_{\text{int}}))$ (uniform additive) & 86.2 & $-$4.6 & 7.8 & 85.9 \\
$0.5 S_{\text{hom}} + 0.3\,\sigma(S_{\text{path}}) + 0.2\,(1{-}P_{\text{int}})$ (weighted additive) & 87.1 & $-$3.7 & 7.3 & 86.4 \\
$\min(S_{\text{hom}},\, \sigma(S_{\text{path}}),\, (1{-}P_{\text{int}}))$ (worst-component) & 88.4 & $-$2.4 & 6.2 & 89.1 \\
$S_{\text{hom}}^{0.5} \cdot \sigma(S_{\text{path}})^{0.3} \cdot (1{-}P_{\text{int}})^{0.2}$ (geometric weighted) & 89.6 & $-$1.2 & 5.6 & 90.2 \\
\bottomrule
\end{tabular}%
}
\end{table}

\noindent The multiplicative default achieves the highest CSS ($90.8\%$) because a near-zero score on any component e.g., a severe drug interaction ($P_{\text{int}} \approx 1$) drives the product toward zero, creating strong gradient signal for the preference pair. Additive variants dilute this effect: a dangerous interaction can be offset by high $S_{\text{hom}}$ and $S_{\text{path}}$, weakening the safety signal. The worst-component ($\min$) function is the second-best, consistent with the safety-gate intuition, but discards magnitude information that the product preserves.

\begin{table}[H]
\centering
\caption{Steepness Parameter ($\lambda$) Sensitivity on CSS}
\label{tab:lambda_sensitivity}
\resizebox{\columnwidth}{!}{%
\begin{tabular}{@{}lccccc@{}}
\toprule
\textbf{$\lambda$ Perturbation} & \textbf{CSS} & \textbf{$\Delta$CSS} & \textbf{HR$\downarrow$} & \textbf{$\Delta$HR} & \textbf{Note} \\
\midrule
Default $\lambda$ values (Table~5) & \textbf{90.8} &   & \textbf{5.1} &   &   \\
All $\lambda \times 0.5$ (halved steepness) & 87.3 & $-$3.5 & 7.4 & $+$2.3 & Softer penalties \\
All $\lambda \times 2.0$ (doubled steepness) & 89.2 & $-$1.6 & 5.9 & $+$0.8 & Harsher but noisier \\
Uniform $\lambda = 2.0$ for all constraints & 88.1 & $-$2.7 & 6.8 & $+$1.7 & Ignores clinical risk profiles \\
$\lambda$ from Gaussian noise ($\pm 20\%$) & 89.9 & $-$0.9 & 5.5 & $+$0.4 & Robust to moderate noise \\
\bottomrule
\end{tabular}%
}
\end{table}

\noindent CSS is moderately sensitive to $\lambda$ choices: halving all steepness values degrades CSS by 3.5~pp because violations near boundaries are under-penalized, while doubling them causes a smaller 1.6~pp drop from over-penalization of borderline-normal values. The constraint-specific defaults derived from clinical risk profiles outperform a uniform $\lambda$, justifying the per-constraint parameterization. Importantly, $\pm 20\%$ Gaussian perturbation of the default $\lambda$ values degrades CSS by only 0.9~pp, indicating that the method is not brittle to moderate uncertainty in the steepness calibration.

\begin{table}[H]
\centering
\caption{Performance Progression Across Training Iterations}
\label{tab:iteration_analysis_appendix}
\resizebox{\columnwidth}{!}{%
\begin{tabular}{lccccc}
\toprule
\textbf{Iteration} & \textbf{MA (\%)} & \textbf{CSS (\%)} & \textbf{HR (\%)$\downarrow$} & \textbf{DID (\%)} & \textbf{PC} \\
\midrule
Iteration 1 & 71.3 & 78.2 & 10.4 & 81.5 & 0.79 \\
Iteration 2 & 72.4 & 82.6 & 8.9 & 84.2 & 0.82 \\
Iteration 3 & 73.8 & 86.4 & 7.2 & 87.3 & 0.85 \\
Iteration 4 & 74.6 & 89.1 & 5.8 & 90.1 & 0.87 \\
Iteration 5 & 75.2 & 90.8 & 5.1 & 91.6 & 0.89 \\
\midrule
$\Delta$ (Iter 1$\rightarrow$5) & +3.9 & +12.6 & -5.3 & +10.1 & +0.10 \\
\bottomrule
\end{tabular}%
}
\end{table}

\begin{table}[H]
\centering
\caption{Cross-Domain Generalization Results}
\label{tab:cross_domain_detailed}
\footnotesize
\setlength{\tabcolsep}{3pt}
\begin{tabular}{@{}l@{\hskip 5pt}c@{\hskip 5pt}c@{\hskip 5pt}c@{\hskip 5pt}c@{\hskip 5pt}c@{}}
\toprule
\textbf{Specialty} & \textbf{CSS} & \textbf{CSS} & \textbf{$\Delta$} & \textbf{HR} & \textbf{DID} \\
 & \textbf{(Ours)} & \textbf{(ORPO)} & & \textbf{(Ours)} & \textbf{(Ours)} \\
\midrule
Cardiology & 91.2 & 70.3 & +20.9 & 4.8 & 92.4 \\
Nephrology & 89.7 & 68.1 & +21.6 & 5.3 & 90.8 \\
Oncology & 87.4 & 65.8 & +21.6 & 6.1 & 88.2 \\
Pediatrics & 86.9 & 64.2 & +22.7 & 5.8 & 89.1 \\
Psychiatry & 84.3 & 67.5 & +16.8 & 7.2 & 82.6 \\
Emerg.\ Med. & 88.6 & 66.4 & +22.2 & 5.5 & 89.7 \\
Geriatrics & 87.1 & 63.8 & +23.3 & 6.4 & 88.4 \\
\midrule
\textbf{Average} & \textbf{87.9} & \textbf{66.6} & \textbf{+21.3} & \textbf{5.9} & \textbf{88.7} \\
\bottomrule
\end{tabular}
\end{table}

\begin{figure}[H]
\centering
\begin{tikzpicture}
\begin{axis}[
    width=0.95\columnwidth,
    height=6cm,
    xlabel={Training Steps ($\times$ 1000)},
    ylabel={Loss},
    xmin=0, xmax=30,
    ymin=0, ymax=2.5,
    legend pos=north east,
    legend style={font=\small, draw=none, fill=none},
    grid=major,
    grid style={dashed, gray!30},
    axis line style={draw=none},
    tick label style={font=\small}
]
\addplot[name path=total, color=nature1, thick, smooth] coordinates {(0,2.3) (5,1.4) (10,0.9) (15,0.6) (20,0.48) (25,0.42) (30,0.38)};
\addplot[name path=sft, color=nature2, thick, smooth] coordinates {(0,2.3) (5,1.5) (10,1.0) (15,0.75) (20,0.65) (25,0.62) (30,0.6)};
\addplot[name path=kg, color=nature3, thick, smooth] coordinates {(0,0.8) (5,0.45) (10,0.28) (15,0.18) (20,0.14) (25,0.12) (30,0.11)};
\addplot[name path=or, color=nature5, thick, dashed, smooth] coordinates {(0,0) (5,0.25) (10,0.35) (15,0.38) (20,0.36) (25,0.32) (30,0.28)};

\path[name path=axis] (axis cs:0,0) -- (axis cs:30,0);
\addplot[color=nature1, fill opacity=0.1] fill between[of=total and axis];

\legend{Total Loss, SFT Loss, KG Loss, OR Loss}
\end{axis}
\end{tikzpicture}
\caption{Training loss convergence over 30K steps. Total loss decreases from 2.3 to 0.38, with KG loss showing the fastest convergence. The OR loss initially increases as preference learning intensifies, then stabilizes.}
\label{fig:training_dynamics_appendix}
\end{figure}

\begin{table}[H]
\centering
\caption{Training Statistics Across Iterations}
\label{tab:training_statistics}
\begin{tabular}{lccccc}
\toprule
\textbf{Metric} & \textbf{Iter 1} & \textbf{Iter 2} & \textbf{Iter 3} & \textbf{Iter 4} & \textbf{Iter 5} \\
\midrule
Total Loss & 0.82 & 0.61 & 0.48 & 0.41 & 0.38 \\
SFT Loss & 0.95 & 0.78 & 0.68 & 0.62 & 0.60 \\
KG Loss & 0.28 & 0.18 & 0.14 & 0.12 & 0.11 \\
OR Loss & 0.35 & 0.32 & 0.30 & 0.29 & 0.28 \\
Pref. Pairs Generated & 1,750 & 1,750 & 1,750 & 1,750 & 1,750 \\
Avg. Score Gap & 0.42 & 0.38 & 0.31 & 0.26 & 0.22 \\
GPU Hours & 3.2 & 3.4 & 3.6 & 3.8 & 4.0 \\
\bottomrule
\end{tabular}
\end{table}

\section{Implementation Details}

We implement our framework using multiple open-source LLMs to ensure reproducibility and broad applicability. Table~\ref{tab:hyperparameters} summarizes all hyperparameters.

\begin{table}[!ht]
\centering
\caption{Hyperparameter Configuration}
\label{tab:hyperparameters}
\resizebox{\columnwidth}{!}{
\begin{tabular}{llc}
\toprule
\textbf{Component} & \textbf{Parameter} & \textbf{Value} \\
\midrule
\multirow{6}{*}{LLM Training} & Base Model & Mistral-7B-v0.3 \\
& Learning Rate & $2 \times 10^{-5}$ \\
& LR Schedule & Cosine Decay \\
& Batch Size & 32 \\
& Max Sequence Length & 2048 \\
& Gradient Accumulation & 4 \\
\midrule
\multirow{4}{*}{LoRA} & Rank ($r$) & 64 \\
& Alpha ($\alpha$) & 128 \\
& Dropout & 0.05 \\
& Target Modules & q\_proj, v\_proj, k\_proj, o\_proj \\
\midrule
\multirow{5}{*}{HGNN} & Hidden Dimension & 256 \\
& Number of Layers & 4 \\
& Attention Heads & 8 \\
& Dropout & 0.1 \\
& Activation & LeakyReLU (0.2) \\
\midrule
\multirow{5}{*}{Iterative ORPO} & Iterations ($T$) & 5 \\
& Samples per Query ($K$) & 8 \\
& Initial $\lambda_0$ & 0.1 \\
& Curriculum $\alpha$ & 0.2 \\
& KG Loss Weight ($\gamma$) & 0.05 \\
\midrule
\multirow{3}{*}{CTS Weights} & $\omega_1$ (Physiological) & 0.4 \\
& $\omega_2$ (Safety) & 0.4 \\
& $\omega_3$ (Coherence) & 0.2 \\
\bottomrule
\end{tabular}
}
\end{table}

\textbf{Computational Resources}: All experiments were conducted on 8$\times$ NVIDIA A100 80GB GPUs with mixed-precision training (bfloat16). We utilize the HuggingFace Transformers library with DeepSpeed ZeRO-3 optimization for distributed training. Table~\ref{tab:compute_resources} details computational requirements.

\begin{table}[!ht]
\centering
\caption{Computational Resource Requirements}
\label{tab:compute_resources}
\resizebox{\columnwidth}{!}{
\begin{tabular}{lccc}
\toprule
\textbf{Stage} & \textbf{GPU Hours} & \textbf{Memory (GB)} & \textbf{Storage (GB)} \\
\midrule
KG Construction & 2.4 & 128 & 45 \\
HGNN Pre-training & 8.6 & 320 & 12 \\
LLM Fine-tuning (per iter.) & 3.2 & 640 & 28 \\
Response Generation & 1.8 & 160 & 8 \\
Preference Scoring & 0.4 & 64 & 2 \\
\midrule
\textbf{Total (5 iterations)} & 42.6 &   & 95 \\
\bottomrule
\end{tabular}
}
\end{table}

\textbf{Reproducibility Details}: For entity extraction, we use scispaCy~v0.5.1 with the \texttt{en\_core\_sci\_lg} model (768-dimensional embeddings), with the \texttt{scispacy-linker} component configured to link entities to UMLS CUIs using a similarity threshold of 0.85 and a maximum of 5 candidate CUIs per entity. Drug node embeddings are initialized from Morgan fingerprints with radius~2 and 2,048-bit vector length, projected to the HGNN hidden dimension ($d=256$) via a learned linear transformation. The CTS weights $\omega_1 = 0.4$, $\omega_2 = 0.4$, $\omega_3 = 0.2$ were selected via grid search over the validation set with step size 0.1, subject to the constraint $\omega_1 + \omega_2 + \omega_3 = 1$. The ``Memory (GB)'' column in Table~\ref{tab:compute_resources} reports \emph{aggregate} memory across all 8 GPUs (i.e., 128~GB total = 16~GB per GPU for KG construction, which includes graph materialization and initial node embedding computation). Preference scoring at 0.4 GPU-hours per iteration evaluates $K=8$ candidates for each of the 1,750 training queries and retains one winner--loser preference pair per query.

\section{Additional Tables}

\begin{table}[!ht]
\centering
\caption{Dataset Statistics and Characteristics}
\label{tab:dataset_statistics}
\begin{tabular}{lcccc}
\toprule
\textbf{Dataset} & \textbf{Train} & \textbf{Valid} & \textbf{Test} & \textbf{Task Type} \\
\midrule
MedQA-USMLE & 10,178 & 1,272 & 1,273 & Multiple Choice \\
PubMedQA & 450 & 50 & 500 & Yes/No/Maybe \\
MedMCQA & 182,822 & 4,183 & 6,150 & Multiple Choice \\
MMLU-Medical & 1,089 & 123 & 1,871 & Multiple Choice \\
CSB (Ours) & 1,750 & 250 & 500 & Open Generation \\
DDI-Corpus & 4,521 & 645 & 1,290 & Interaction Detection \\
i2b2-2010 & 16,315 & 2,330 & 4,661 & Relation Extraction \\
\midrule
\textbf{Total} & 217,125 & 8,853 & 16,245 &   \\
\bottomrule
\end{tabular}
\end{table}

\begin{table}[!ht]
\centering
\caption{Clinical Safety Benchmark (CSB) Composition}
\label{tab:csb_breakdown}
\begin{tabular}{lcc}
\toprule
\textbf{Category} & \textbf{Scenarios} & \textbf{Percentage} \\
\midrule
Drug-Drug Interactions & 625 & 25.0\% \\
Contraindication Detection & 500 & 20.0\% \\
Homeostatic Violations & 450 & 18.0\% \\
Dosing Errors (Pediatric/Geriatric) & 375 & 15.0\% \\
Organ Function Impairment & 300 & 12.0\% \\
Multi-system Polypharmacy & 250 & 10.0\% \\
\midrule
\textbf{Total} & 2,500 & 100\% \\
\bottomrule
\end{tabular}
\end{table}

\begin{table}[!ht]
\centering
\caption{Performance Across Different Base Models}
\label{tab:model_scale}
\resizebox{\columnwidth}{!}{
\begin{tabular}{lcccc}
\toprule
\textbf{Model} & \textbf{Params} & \textbf{CSS} & \textbf{HR$\downarrow$} & \textbf{Time} \\
\midrule
BioMistral-7B & 7B & 88.4 & 6.2 & 14.2h \\
Mistral-7B & 7B & 90.8 & 5.1 & 18.6h \\
LLaMA-3-8B & 8B & 91.2 & 4.9 & 19.8h \\
Qwen2-7B & 7B & 89.6 & 5.4 & 17.4h \\
\bottomrule
\end{tabular}
}
\end{table}

\begin{table}[!ht]
\centering
\caption{Taxonomy of Failure and Root Cause Analysis (training+validation scenarios)}
\label{tab:error_analysis}
\resizebox{\columnwidth}{!}{
\begin{tabular}{p{2.9cm}ccp{4.8cm}p{4.3cm}}
\toprule
\textbf{Error Type} & \textbf{Count} & \textbf{\%} & \textbf{Primary Root Cause} & \textbf{Potential Mitigation} \\
\midrule
Graph Retrieval Failures & 84 & 4.2 & Knowledge graph coverage gaps / anatomical mismatch & Dynamic KG updates and continual HGNN refresh \\
Optimization Drift & 62 & 3.1 & Constraint boundary deviation in HGNN depth & Stronger KL regularization and score-gap calibration \\
Linguistic Hallucinations & 56 & 2.8 & Temporal fabrication outside graph representation & Temporal constraints and post-hoc consistency checks \\
Dosing Errors & 41 & 2.1 & Incomplete pharmacokinetic data & Weight/renal-aware dosing calculators in tool loop \\
Context Ambiguity & 38 & 1.9 & Insufficient patient information & Uncertainty-aware abstention and clarification prompts \\
Novel Drug Combinations & 29 & 1.5 & Emerging pharmacological knowledge & Rapid ingestion of approvals and interaction bulletins \\
Cultural/Genetic Factors & 22 & 1.1 & Population-specific variations & Pharmacogenomic personalization and subgroup calibration \\
\midrule
\textbf{Total Errors} & 332 & 16.7 &   &   \\
\bottomrule
\end{tabular}
}
\end{table}

\begin{figure*}[!htb]
\centering
\begin{tikzpicture}[scale=0.93, transform shape, >=Stealth, font=\normalsize,
  stepbox/.style={draw=black!50, fill=white, rectangle,
              rounded corners=3pt, minimum width=5.6cm,
              minimum height=5.2cm, inner sep=0pt, line width=0.5pt},
  badge/.style={circle, fill=nature1, text=white,
              font=\normalsize\bfseries, inner sep=0pt,
              minimum size=18pt},
  arr/.style={-{Stealth[length=5pt,width=4pt]},
              line width=0.8pt, nature1!70},
  arrlbl/.style={font=\footnotesize\sffamily, text=naturegray,
              fill=white, inner sep=2pt, rounded corners=1pt},
  gnode/.style={circle, draw=black!55, inner sep=0pt,
              minimum size=22pt, font=\footnotesize\sffamily,
              line width=0.45pt},
  gedge/.style={draw=black!45, line width=0.4pt, -{Stealth[length=3pt]}},
]

\node[stepbox] (b1) at (0,0) {};
\begin{scope}
  \clip[rounded corners=3pt] (b1.north west) rectangle
       ([yshift=-0.75cm]b1.north east);
  \fill[nature1!10] (b1.north west) rectangle
       ([yshift=-0.75cm]b1.north east);
\end{scope}
\draw[black!50, line width=0.5pt, rounded corners=3pt]
     (b1.south west) rectangle (b1.north east);
\node[badge] at ([xshift=10pt, yshift=-0.38cm]b1.north west) {1};
\node[font=\footnotesize\sffamily\bfseries, anchor=west]
     at ([xshift=24pt, yshift=-0.38cm]b1.north west)
     {Generate Candidate Plan};

\node[anchor=north west, align=left, inner sep=6pt,
      text width=5.2cm] at ([yshift=-0.82cm]b1.north west) {%
  \smallskip
  {\footnotesize\bfseries Patient context}\\[3pt]
  {\footnotesize
    \textbullet~Chronic kidney disease (CKD)\\
    \textbullet~Hyperkalemia:\; $K^{+}=5.4$\,mEq/L\\
    \textbullet~Renal function:\; eGFR\,$=25$}\\[6pt]
  {\footnotesize\bfseries LLM candidate output}\\[2pt]
  \colorbox{nature4!12}{\parbox{4.9cm}{\footnotesize\itshape
   ``Prescribe Spironolactone\,+\\
   Lisinopril for BP control.''}}
};

\node[stepbox] (b2) at (6.6,0) {};
\begin{scope}
  \clip[rounded corners=3pt] (b2.north west) rectangle
       ([yshift=-0.75cm]b2.north east);
  \fill[nature1!10] (b2.north west) rectangle
       ([yshift=-0.75cm]b2.north east);
\end{scope}
\draw[black!50, line width=0.5pt, rounded corners=3pt]
     (b2.south west) rectangle (b2.north east);
\node[badge] at ([xshift=10pt, yshift=-0.38cm]b2.north west) {2};
\node[font=\footnotesize\sffamily\bfseries, anchor=west]
     at ([xshift=24pt, yshift=-0.38cm]b2.north west)
     {Entity Linking \& Sub-graph};

\node[font=\scriptsize, text=naturegray, anchor=north west]
     at ([xshift=5pt, yshift=-0.9cm]b2.north west)
     {scispaCy NER\;\,$\rightarrow$\;\,UMLS / KG nodes};

\begin{scope}[shift={([yshift=-0.15cm]b2.center)}]
  \node[gnode, fill=nature5!15] (nSpiro) at (-1.7, 0.65) {Spiro};
  \node[gnode, fill=nature5!15] (nLisin) at (-1.7,-0.75) {Lisin};
  \node[gnode, fill=nature4!20] (nK)     at ( 0.6, 0.65) {K\textsuperscript{+}};
  \node[gnode, fill=nature2!12] (nCKD)   at ( 0.6,-0.75) {CKD};
  \node[gnode, fill=nature3!15] (neGFR)  at ( 2.2,-0.05) {eGFR};
  \draw[gedge] (nSpiro) -- node[above=2pt, font=\scriptsize, text=black!55, fill=white, inner sep=1pt]{$\uparrow$K\textsuperscript{+}} (nK);
  \draw[gedge] (nLisin) -- node[below=2pt, font=\scriptsize, text=black!55, fill=white, sloped, inner sep=1pt]{$\uparrow$K\textsuperscript{+}} (nK);
  \draw[gedge] (nCKD)   -- node[above=2pt, font=\scriptsize, text=black!55, fill=white, inner sep=1pt, sloped]{impairs} (neGFR);
  \draw[gedge, <->] (nK) -- node[left=3pt, font=\scriptsize, text=black!55, fill=white, inner sep=1pt]{risk} (nCKD);
\end{scope}

\node[anchor=south west, font=\scriptsize, text=naturegray,
      inner sep=3pt] at ([xshift=2pt, yshift=2pt]b2.south west) {%
  \textcolor{nature5}{$\blacksquare$}\,Drug\;
  \textcolor{nature4!80!black}{$\blacksquare$}\,Lab\;
  \textcolor{nature2!70!black}{$\blacksquare$}\,Cond\;
  \textcolor{nature3}{$\blacksquare$}\,Organ};

\node[stepbox] (b3) at (13.2,0) {};
\begin{scope}
  \clip[rounded corners=3pt] (b3.north west) rectangle
       ([yshift=-0.75cm]b3.north east);
  \fill[nature2!8] (b3.north west) rectangle
       ([yshift=-0.75cm]b3.north east);
\end{scope}
\draw[black!50, line width=0.5pt, rounded corners=3pt]
     (b3.south west) rectangle (b3.north east);
\node[badge, fill=nature2] at ([xshift=10pt, yshift=-0.38cm]b3.north west) {3};
\node[font=\footnotesize\sffamily\bfseries, anchor=west]
     at ([xshift=24pt, yshift=-0.38cm]b3.north west)
     {HGNN Constraint Check};

\node[anchor=north west, align=left, inner sep=6pt,
      text width=5.2cm] (checks) at ([yshift=-0.82cm]b3.north west) {%
  \smallskip
  {\footnotesize\bfseries Constraint violations}\\[3pt]
  {\footnotesize
    \textcolor{nature2}{\ding{55}}\;$K^{+}$ boundary
      $\mathcal{B}=[3.5,\,5.0]$ exceeded\\[2pt]
    \textcolor{nature2}{\ding{55}}\;ACEi\,+\,CKD\,+\,high
      $K^{+}$ risk\\[2pt]
    \textcolor{nature2}{\ding{55}}\;K-sparing diuretic
      contraindicated}\\[8pt]
  {\footnotesize\bfseries PhysioScore}\\[2pt]
  {\footnotesize $s=0.02$\;\;%
   (threshold $\tau=0.5$)\;\;%
   \textcolor{nature2}{\textbf{FAIL}}}
};

\node[draw=nature2, fill=nature2!6, line width=0.8pt,
      rounded corners=3pt, inner xsep=10pt, inner ysep=5pt,
      font=\footnotesize\sffamily\bfseries,
      minimum width=10cm, anchor=north]
  (verdict) at ([yshift=-8pt]b2.south)
  {\textcolor{nature2}{\ding{55}}\;\;Verdict:\;%
   \textsc{Unsafe}\; \;Plan Rejected\;\;%
   {\normalfont\scriptsize(3/3 constraints violated)}};

\draw[arr] ([yshift=2pt]b1.east) -- ([yshift=2pt]b2.west)
  node[arrlbl, midway, above=1pt] {entities};
\draw[arr] ([yshift=2pt]b2.east) -- ([yshift=2pt]b3.west)
  node[arrlbl, midway, above=1pt] {sub-graph};
\draw[arr, nature2!55] (b3.south) -- (b3.south |- verdict.east)
  -- (verdict.east);

\node[font=\scriptsize, text=black!45, anchor=north]
  at ([yshift=-5pt]verdict.south) {%
  Follows Algorithm~\ref{alg:hgnn_scoring}:\;%
  \textsc{EntityExtract}\;$\rightarrow$\;%
  \textsc{MapToGraph}\;$\rightarrow$\;%
  \textsc{PhysioScore}};

\end{tikzpicture}
\caption{Step-wise toy example of constraint verification. A candidate plan
is generated, entities are linked to the physiological graph, and HGNN-based
checks reject the plan due to hyperkalemia-related contraindications.}
\label{fig:toy_example}
\end{figure*}

\begin{table}[H]
\centering
\caption{Human Evaluation Rubric}
\label{tab:human_eval_rubric}
\resizebox{\columnwidth}{!}{%
\begin{tabular}{cp{14cm}}
\toprule
\textbf{Score} & \textbf{Criteria} \\
\midrule
5 & Clinically excellent: Safe, accurate, complete, follows guidelines \\
4 & Clinically acceptable: Minor omissions, no safety concerns \\
3 & Borderline: Some inaccuracies, requires verification \\
2 & Concerning: Contains errors that could mislead \\
1 & Dangerous: Contains recommendations that could cause harm \\
\bottomrule
\end{tabular}%
}
\end{table}

\begin{table}[H]
\centering
\caption{Evaluation Metrics Overview}
\label{tab:metrics_overview}
\resizebox{\columnwidth}{!}{%
\begin{tabular}{lp{12cm}}
\toprule
\textbf{Metric} & \textbf{Description} \\
\midrule
MA & Medical Accuracy: Percentage correct on QA benchmarks \\
CSS & Clinical Safety Score: Adherence to physiological constraints \\
HR & Hallucination Rate: \% responses with impossible statements \\
DID & Drug Interaction Detection: F1 for contraindicated combinations \\
PC & Physiological Consistency: Mean HGNN feasibility score \\
BLEU-Med & Modified BLEU with medical term weighting \\
ROUGE-L & Longest common subsequence for generation quality \\
BertScore & Semantic similarity using BioBERT embeddings \\
\bottomrule
\end{tabular}%
}
\end{table}

\end{document}

%% file: graphical_abstract_tikz.tex
\begingroup
\linespread{1.15}\selectfont
\let\origscriptsize\scriptsize
\let\origfootnotesize\footnotesize
\let\origsmall\small
\let\orignormalsize\normalsize
\let\origlarge\large
\let\origLarge\Large
\renewcommand{\scriptsize}{\origsmall}
\renewcommand{\footnotesize}{\orignormalsize}
\renewcommand{\small}{\origlarge}
\renewcommand{\normalsize}{\origLarge}
\begin{tikzpicture}[
    scale=0.65,
    transform shape,
    every node/.style={font=\sffamily\Huge},
]

\definecolor{tealheader}{RGB}{0,102,102}
\definecolor{beigeback}{RGB}{255,250,240}
\definecolor{purplefooter}{RGB}{130,90,170}
\definecolor{orangeaccent}{RGB}{220,120,40}
\definecolor{greenaccent}{RGB}{40,160,80}
\definecolor{blueaccent}{RGB}{50,120,190}
\definecolor{redaccent}{RGB}{200,60,60}
\definecolor{graytext}{RGB}{50,50,50}
\definecolor{lightgray}{RGB}{250,250,250}
\definecolor{darkgray}{RGB}{35,35,35}
\definecolor{tanback}{RGB}{255,248,235}
\definecolor{cyanlight}{RGB}{230,250,250}

\fill[white] (-1,-9.5) rectangle (30,10.5);


\begin{scope}[shift={(0,5.5)}]
    \fill[lightgray, rounded corners=4pt] (0,0) rectangle (4,3);
    \draw[graytext!60, line width=1pt, rounded corners=4pt] (0,0) rectangle (4,3);
    
    \fill[white, rounded corners=3pt] (0.6,0.5) rectangle (3.4,2.5);
    \draw[graytext!70, line width=0.8pt, rounded corners=3pt] (0.6,0.5) rectangle (3.4,2.5);
    \foreach \y in {2.1,1.7,1.3,0.9} {
        \draw[graytext!40, line width=0.6pt] (0.9,\y) -- (3.1,\y);
    }
    \node[font=\normalsize\bfseries, text=tealheader] at (2,0.65) {[query]};
    
    \node[font=\normalsize\bfseries, text=darkgray, align=center] at (2,-0.6) {Clinical Query\\(EHR Context)};
\end{scope}

\draw[-{Stealth[length=5mm,width=4mm]}, line width=2.5pt, color=darkgray!70] (4.3,7) -- (5.8,7);

\begin{scope}[shift={(6.2,4.5)}]
    \fill[cyanlight, rounded corners=6pt] (0,0) rectangle (11,4.5);
    \draw[tealheader, line width=2pt, rounded corners=6pt] (0,0) rectangle (11,4.5);
    
    \node[font=\Large\bfseries, text=darkgray] at (5.5,3.9) {Neuro-Symbolic Alignment Framework};
    
    \fill[white, rounded corners=4pt] (0.5,2.4) rectangle (10.5,3.3);
    \draw[tealheader!60, line width=1pt, rounded corners=4pt] (0.5,2.4) rectangle (10.5,3.3);
    \node[font=\normalsize\bfseries, text=darkgray] at (5.5,2.85) {LLM Response Generation \& Candidate Sampling};
    
    \fill[white, rounded corners=4pt] (0.5,1.3) rectangle (10.5,2.2);
    \draw[tealheader!60, line width=1pt, rounded corners=4pt] (0.5,1.3) rectangle (10.5,2.2);
    \node[font=\normalsize\bfseries, text=darkgray] at (5.5,1.75) {HGNN Physiological Verification};
    
    \fill[white, rounded corners=4pt] (0.5,0.2) rectangle (10.5,1.1);
    \draw[tealheader!60, line width=1pt, rounded corners=4pt] (0.5,0.2) rectangle (10.5,1.1);
    \node[font=\normalsize\bfseries, text=darkgray] at (5.5,0.65) {Iterative ORPO Alignment};
\end{scope}

\draw[-{Stealth[length=5mm,width=4mm]}, line width=2.5pt, color=darkgray!70] (17.5,7) -- (19,7);

\begin{scope}[shift={(19.5,4.5)}]
    \fill[lightgray, rounded corners=4pt] (0,0) rectangle (7.5,4.5);
    \draw[graytext!60, line width=1pt, rounded corners=4pt] (0,0) rectangle (7.5,4.5);
    
    \fill[graytext!25, rounded corners=4pt] (0.3,3.5) rectangle (7.2,4.2);
    \node[font=\normalsize\bfseries, text=darkgray] at (1.0,3.85) {Query};
    \node[font=\normalsize\bfseries, text=darkgray] at (2.6,3.85) {Response};
    \node[font=\normalsize\bfseries, text=darkgray] at (4.3,3.85) {Safety};
    \node[font=\normalsize\bfseries, text=darkgray] at (5.7,3.85) {Score};
    \node[font=\normalsize\bfseries, text=darkgray] at (6.8,3.85) {KG};
    
    \node[font=\small\bfseries, text=darkgray] at (1.0,3.0) {[1]};
    \node[font=\small\bfseries, text=greenaccent] at (2.6,3.0) {Safe};
    \node[fill=greenaccent!40, rounded corners=3pt, inner sep=4pt, font=\small\bfseries, text=greenaccent!90!black] at (4.3,3.0) {Pass};
    \node[font=\small\bfseries, text=darkgray] at (5.7,3.0) {0.92};
    \fill[greenaccent!50] (6.4,2.9) rectangle (7.1,3.1);
    
    \node[font=\small\bfseries, text=darkgray] at (1.0,2.4) {[2]};
    \node[font=\small\bfseries, text=redaccent] at (2.6,2.4) {Unsafe};
    \node[fill=redaccent!40, rounded corners=3pt, inner sep=4pt, font=\small\bfseries, text=redaccent!90!black] at (4.3,2.4) {Fail};
    \node[font=\small\bfseries, text=darkgray] at (5.7,2.4) {0.31};
    \fill[redaccent!50] (6.4,2.3) rectangle (6.7,2.5);
    
    \node[font=\small\bfseries, text=darkgray] at (1.0,1.8) {[3]};
    \node[font=\small\bfseries, text=greenaccent] at (2.6,1.8) {Safe};
    \node[fill=greenaccent!40, rounded corners=3pt, inner sep=4pt, font=\small\bfseries, text=greenaccent!90!black] at (4.3,1.8) {Pass};
    \node[font=\small\bfseries, text=darkgray] at (5.7,1.8) {0.89};
    \fill[greenaccent!50] (6.4,1.7) rectangle (7.0,1.9);
    
    \fill[tanback, rounded corners=3pt] (0.4,0.3) rectangle (7.1,1.3);
    \draw[orangeaccent, line width=1pt, rounded corners=3pt] (0.4,0.3) rectangle (7.1,1.3);
    \node[font=\normalsize\bfseries, text=darkgray] at (3.75,0.95) {Clinical Safety Summary};
    \node[font=\small\bfseries, text=greenaccent] at (2.2,0.55) {CSS: +34\%};
    \node[font=\small\bfseries, text=redaccent] at (5.3,0.55) {Halluc: -64\%};
\end{scope}


\begin{scope}[shift={(0,-1)}]
    \fill[tanback, rounded corners=6pt] (0,0) rectangle (8,4.5);
    \draw[orangeaccent, line width=1.5pt, dashed, rounded corners=6pt] (0,0) rectangle (8,4.5);
    
    \node[font=\Large\bfseries, text=darkgray] at (4,4) {LLM Generation};
    \node[font=\normalsize, text=graytext] at (4,3.5) {Response Candidate Sampling};
    
    \fill[white, rounded corners=4pt] (0.5,1.6) rectangle (3.5,3.1);
    \draw[orangeaccent!60, line width=0.8pt, rounded corners=4pt] (0.5,1.6) rectangle (3.5,3.1);
    
    \foreach \x in {0.7,1.3,1.9,2.5,3.1} {
        \fill[orangeaccent!40, rounded corners=2pt] (\x,1.8) rectangle (\x+0.4,2.9);
    }
    \node[font=\small\bfseries, text=darkgray] at (2,1.35) {$\mathcal{M}_\theta$};
    
    \draw[darkgray!50, line width=0.8pt] (3.5,2.35) -- (4.2,2.8);
    \draw[darkgray!50, line width=0.8pt] (3.5,2.35) -- (4.2,2.35);
    \draw[darkgray!50, line width=0.8pt] (3.5,2.35) -- (4.2,1.9);
    
    \node[fill=orangeaccent!30, rounded corners=3pt, inner sep=5pt, font=\small\bfseries, text=orangeaccent!90!black] at (5.8,2.8) {$r_1$: Ibuprofen};
    \node[fill=orangeaccent!30, rounded corners=3pt, inner sep=5pt, font=\small\bfseries, text=orangeaccent!90!black] at (6.0,2.35) {$r_2$: Acetaminophen};
    \node[fill=orangeaccent!30, rounded corners=3pt, inner sep=5pt, font=\small\bfseries, text=orangeaccent!90!black] at (5.8,1.9) {$r_K$: Tramadol};
    
    \fill[white, rounded corners=3pt] (0.4,0.3) rectangle (7.6,1.1);
    \draw[orangeaccent, line width=1pt, rounded corners=3pt] (0.4,0.3) rectangle (7.6,1.1);
    \node[font=\small\bfseries, text=orangeaccent!90!black] at (4,0.7) {? Which is physiologically safe?};
\end{scope}

\draw[-{Stealth[length=4mm,width=3mm]}, line width=2pt, color=darkgray!60] (8.3,1.25) -- (9.4,1.25);

\begin{scope}[shift={(9.7,-1)}]
    \fill[cyanlight, rounded corners=6pt] (0,0) rectangle (9,4.5);
    \draw[tealheader, line width=1.5pt, dashed, rounded corners=6pt] (0,0) rectangle (9,4.5);
    
    \node[font=\Large\bfseries, text=darkgray] at (4.5,4) {Physiological World Model};
    \node[font=\normalsize, text=graytext] at (4.5,3.5) {Heterogeneous Knowledge Graph};
    
    \fill[white, rounded corners=4pt] (0.5,0.9) rectangle (5,3.1);
    \draw[tealheader!50, line width=0.8pt, rounded corners=4pt] (0.5,0.9) rectangle (5,3.1);
    
    \node[circle, fill=redaccent!70, inner sep=5pt] (d1) at (1.1,2.6) {};
    \node[circle, fill=tealheader!70, inner sep=5pt] (d2) at (2.0,2.8) {};
    \node[circle, fill=orangeaccent!80, inner sep=5pt] (d3) at (3.0,2.4) {};
    \node[circle, fill=blueaccent!70, inner sep=5pt] (d4) at (3.9,2.7) {};
    \node[circle, fill=greenaccent!70, inner sep=5pt] (d5) at (2.5,1.5) {};
    \node[circle, fill=redaccent!50, inner sep=5pt] (d6) at (4.3,1.9) {};
    \node[circle, fill=purplefooter, inner sep=4pt] (d7) at (1.3,1.3) {};
    
    \draw[graytext!50, line width=1pt] (d1)--(d2)--(d3)--(d4);
    \draw[graytext!50, line width=1pt] (d2)--(d5)--(d3);
    \draw[graytext!50, line width=1pt] (d3)--(d6);
    \draw[graytext!50, line width=1pt] (d1)--(d7)--(d5);
    \draw[graytext!50, line width=1pt] (d4)--(d6);
    
    \fill[white, rounded corners=4pt] (5.2,1.6) rectangle (8.7,3.1);
    \draw[tealheader!40, line width=0.6pt, rounded corners=4pt] (5.2,1.6) rectangle (8.7,3.1);
    \node[fill=redaccent!35, rounded corners=3pt, inner sep=3pt, font=\footnotesize\bfseries, text=redaccent!90!black] at (5.8,2.75) {Drug};
    \node[fill=tealheader!35, rounded corners=3pt, inner sep=3pt, font=\footnotesize\bfseries, text=tealheader!90!black] at (6.95,2.75) {Organ};
    \node[fill=orangeaccent!35, rounded corners=3pt, inner sep=3pt, font=\footnotesize\bfseries, text=orangeaccent!90!black] at (8.1,2.75) {Path};
    \node[fill=blueaccent!35, rounded corners=3pt, inner sep=3pt, font=\footnotesize\bfseries, text=blueaccent!90!black] at (6.0,2.2) {Symptom};
    \node[fill=greenaccent!35, rounded corners=3pt, inner sep=3pt, font=\footnotesize\bfseries, text=greenaccent!90!black] at (7.6,2.2) {Disease};
    
    \node[font=\small\bfseries, text=darkgray] at (2.75,0.6) {\textbf{847K} nodes};
    \node[font=\small\bfseries, text=darkgray] at (2.75,0.25) {\textbf{3.2M} edges};
    
    \fill[white, rounded corners=3pt] (5.3,0.2) rectangle (8.6,1.45);
    \draw[tealheader!40, line width=0.6pt, rounded corners=3pt] (5.3,0.2) rectangle (8.6,1.45);
    \node[font=\small\bfseries, text=darkgray] at (6.95,1.2) {Relations:};
    \node[font=\scriptsize, text=graytext] at (6.95,0.85) {regulates, inhibits};
    \node[font=\scriptsize, text=graytext] at (6.95,0.5) {contraindicated};
\end{scope}

\draw[-{Stealth[length=4mm,width=3mm]}, line width=2pt, color=darkgray!60] (19,1.25) -- (20.1,1.25);

\begin{scope}[shift={(20.4,-1)}]
    \fill[purplefooter!25, rounded corners=6pt] (0,0) rectangle (7,4.5);
    \draw[purplefooter, line width=1.5pt, dashed, rounded corners=6pt] (0,0) rectangle (7,4.5);
    
    \node[font=\Large\bfseries, text=darkgray] at (3.5,4) {HGNN Verification};
    \node[font=\normalsize, text=graytext] at (3.5,3.5) {Physiological Feasibility Scoring};
    
    \fill[white, rounded corners=4pt] (0.4,2.2) rectangle (6.6,3.1);
    \draw[purplefooter!60, line width=0.8pt, rounded corners=4pt] (0.4,2.2) rectangle (6.6,3.1);
    \foreach \x in {0.6,1.5,2.4,3.3,4.2,5.1,6.0} {
        \fill[purplefooter!50, rounded corners=2pt] (\x,2.35) rectangle (\x+0.6,2.95);
    }
    \node[font=\small\bfseries, text=darkgray] at (3.5,1.9) {4-layer R-GAT (256-dim)};
    
    \fill[greenaccent!30, rounded corners=3pt] (0.4,1.1) rectangle (2.2,1.6);
    \node[font=\small\bfseries, text=greenaccent!90!black] at (1.3,1.35) {$S_{hom}$: 0.91};
    
    \fill[blueaccent!30, rounded corners=3pt] (2.5,1.1) rectangle (4.3,1.6);
    \node[font=\small\bfseries, text=blueaccent!90!black] at (3.4,1.35) {$S_{path}$: 0.87};
    
    \fill[redaccent!30, rounded corners=3pt] (4.6,1.1) rectangle (6.4,1.6);
    \node[font=\small\bfseries, text=redaccent!90!black] at (5.5,1.35) {$P_{int}$: 0.05};
    
    \fill[greenaccent!40, rounded corners=4pt] (1.2,0.25) rectangle (5.8,0.85);
    \draw[greenaccent, line width=1.2pt, rounded corners=4pt] (1.2,0.25) rectangle (5.8,0.85);
    \node[font=\normalsize\bfseries, text=greenaccent!90!black] at (3.5,0.55) {\ding{51} Final Score: 0.89};
\end{scope}

\begin{scope}[shift={(0,-6)}]
    \fill[tanback, rounded corners=6pt] (0,0) rectangle (27,4);
    \draw[orangeaccent, line width=1.5pt, rounded corners=6pt] (0,0) rectangle (27,4);
    
    \node[font=\Large\bfseries, text=darkgray] at (13.5,3.5) {Iterative ORPO Alignment Process};
    
    \fill[orangeaccent!50, rounded corners=4pt] (1.0,1.4) rectangle (4.5,2.9);
    \draw[orangeaccent!70, line width=0.8pt, rounded corners=4pt] (1.0,1.4) rectangle (4.5,2.9);
    \node[font=\normalsize\bfseries, text=white] at (2.75,2.4) {T1};
    \node[font=\small\bfseries, text=white] at (2.75,1.85) {Init};
    
    \fill[orangeaccent!65, rounded corners=4pt] (5.5,1.4) rectangle (9.0,2.9);
    \draw[orangeaccent!80, line width=0.8pt, rounded corners=4pt] (5.5,1.4) rectangle (9.0,2.9);
    \node[font=\normalsize\bfseries, text=white] at (7.25,2.4) {T2};
    \node[font=\small\bfseries, text=white] at (7.25,1.85) {Gen};
    
    \fill[orangeaccent!80, rounded corners=4pt] (10.0,1.4) rectangle (13.5,2.9);
    \draw[orangeaccent!90, line width=0.8pt, rounded corners=4pt] (10.0,1.4) rectangle (13.5,2.9);
    \node[font=\normalsize\bfseries, text=white] at (11.75,2.4) {T3};
    \node[font=\small\bfseries, text=white] at (11.75,1.85) {Score};
    
    \fill[orangeaccent, rounded corners=4pt] (14.5,1.4) rectangle (18.0,2.9);
    \draw[orangeaccent!90, line width=0.8pt, rounded corners=4pt] (14.5,1.4) rectangle (18.0,2.9);
    \node[font=\normalsize\bfseries, text=white] at (16.25,2.4) {T4};
    \node[font=\small\bfseries, text=white] at (16.25,1.85) {Rank};
    
    \fill[orangeaccent!80!black, rounded corners=4pt] (19.0,1.4) rectangle (22.5,2.9);
    \draw[orangeaccent!90, line width=0.8pt, rounded corners=4pt] (19.0,1.4) rectangle (22.5,2.9);
    \node[font=\normalsize\bfseries, text=white] at (20.75,2.4) {T5};
    \node[font=\small\bfseries, text=white] at (20.75,1.85) {Update};
    
    \draw[-{Stealth[length=2.5mm,width=2mm]}, line width=1.2pt, color=darkgray!60] (4.7,2.15) -- (5.3,2.15);
    \draw[-{Stealth[length=2.5mm,width=2mm]}, line width=1.2pt, color=darkgray!60] (9.2,2.15) -- (9.8,2.15);
    \draw[-{Stealth[length=2.5mm,width=2mm]}, line width=1.2pt, color=darkgray!60] (13.7,2.15) -- (14.3,2.15);
    \draw[-{Stealth[length=2.5mm,width=2mm]}, line width=1.2pt, color=darkgray!60] (18.2,2.15) -- (18.8,2.15);
    
    \node[fill=orangeaccent!30, rounded corners=3pt, inner sep=5pt, font=\small\bfseries, text=orangeaccent!90!black] at (3.5,0.65) {$\mathcal{L}_{SFT}$};
    \node[font=\footnotesize\bfseries, text=darkgray] at (3.5,0.2) {Supervised};
    
    \node[fill=orangeaccent!50, rounded corners=3pt, inner sep=5pt, font=\small\bfseries, text=orangeaccent!90!black] at (13.5,0.65) {$\mathcal{L}_{OR}$};
    \node[font=\footnotesize\bfseries, text=darkgray] at (13.5,0.2) {Odds Ratio};
    
    \node[fill=orangeaccent!70, rounded corners=3pt, inner sep=5pt, font=\small\bfseries, text=white] at (23.5,0.65) {$\mathcal{L}_{KG}$};
    \node[font=\footnotesize\bfseries, text=darkgray] at (23.5,0.2) {KG Consistency};
\end{scope}

\begin{scope}[shift={(0,-9)}]
    \fill[purplefooter!35, rounded corners=5pt] (0,0) rectangle (27,2.6);
    \draw[purplefooter, line width=1.2pt, rounded corners=5pt] (0,0) rectangle (27,2.6);
    
    \node[font=\Large\bfseries, text=darkgray] at (13.5,2.2) {Key Components \& Metrics};
    
    \fill[blueaccent!20, rounded corners=4pt] (0.4,0.35) rectangle (5.2,1.8);
    \draw[blueaccent!60, line width=0.8pt, rounded corners=4pt] (0.4,0.35) rectangle (5.2,1.8);
    \node[font=\normalsize\bfseries, text=blueaccent!90!black] at (2.8,1.45) {LLM -- Neural};
    \node[font=\small, text=darkgray, text width=4.2cm, align=center] at (2.8,0.85) {Response generation,\\candidate sampling};
    
    \fill[tealheader!20, rounded corners=4pt] (5.6,0.35) rectangle (10.8,1.8);
    \draw[tealheader!60, line width=0.8pt, rounded corners=4pt] (5.6,0.35) rectangle (10.8,1.8);
    \node[font=\normalsize\bfseries, text=tealheader!90!black] at (8.2,1.45) {KG -- Symbolic};
    \node[font=\small, text=darkgray, text width=4.6cm, align=center] at (8.2,0.85) {847K nodes, 3.2M edges\\physiological constraints};
    
    \fill[purplefooter!40, rounded corners=4pt] (11.2,0.35) rectangle (16.4,1.8);
    \draw[purplefooter, line width=0.8pt, rounded corners=4pt] (11.2,0.35) rectangle (16.4,1.8);
    \node[font=\normalsize\bfseries, text=purplefooter!90!black] at (13.8,1.45) {HGNN -- Verify};
    \node[font=\small, text=darkgray, text width=4.6cm, align=center] at (13.8,0.85) {4-layer R-GAT\\feasibility scoring};
    
    \fill[orangeaccent!25, rounded corners=4pt] (16.8,0.35) rectangle (22.0,1.8);
    \draw[orangeaccent!70, line width=0.8pt, rounded corners=4pt] (16.8,0.35) rectangle (22.0,1.8);
    \node[font=\normalsize\bfseries, text=orangeaccent!90!black] at (19.4,1.45) {ORPO -- Align};
    \node[font=\small, text=darkgray, text width=4.6cm, align=center] at (19.4,0.85) {Iterative preference\\optimization (T=5)};
    
    \fill[greenaccent!25, rounded corners=4pt] (22.4,0.35) rectangle (26.6,1.8);
    \draw[greenaccent!70, line width=0.8pt, rounded corners=4pt] (22.4,0.35) rectangle (26.6,1.8);
    \node[font=\normalsize\bfseries, text=greenaccent!90!black] at (24.5,1.45) {Results};
    \node[font=\small, text=darkgray, text width=3.8cm, align=center] at (24.5,0.85) {+34\% CSS, -64\% HR\\91.6\% DID, 0.89 PC};
\end{scope}

\end{tikzpicture}
\endgroup

%% file: pipeline_scaled.tex

\definecolor{stageAred}{RGB}{205,60,60}
\definecolor{stageBblue}{RGB}{45,90,165}
\definecolor{stageCpurp}{RGB}{120,60,165}
\definecolor{outgreen}{RGB}{45,140,70}
\definecolor{outgreenfill}{RGB}{230,245,230}
\definecolor{datagray}{RGB}{100,105,110}

\begin{figure*}[!t]
\centering
\resizebox{\linewidth}{!}{%
\begin{tikzpicture}[
  every node/.style={font=\fontsize{20}{24}\selectfont},
  >=Stealth,
  ibox/.style={
    rectangle, rounded corners=7.2pt,
    draw=#1!80!black, fill=#1,
    line width=1.4pt,
    minimum height=1.5cm, minimum width=5.0cm,
    align=center, inner sep=7.2pt,
    font=\fontsize{20}{24}\selectfont\bfseries, text=white
  },
  iboxlt/.style={
    rectangle, rounded corners=7.2pt,
    draw=#1!60!black, fill=#1!12,
    line width=1.4pt,
    minimum height=1.5cm, minimum width=5.0cm,
    align=center, inner sep=7.2pt,
    font=\fontsize{20}{24}\selectfont\bfseries, text=#1!80!black
  },
  grp/.style={
    rectangle, rounded corners=10.8pt,
    draw=#1, line width=2.5pt, dashed,
    inner sep=32.4pt, fill=none
  },
  dnode/.style={
    ellipse, draw=#1!80!black, fill=#1!12,
    line width=1.4pt,
    minimum width=4.0cm, minimum height=1.8cm,
    align=center, inner sep=5.4pt,
    font=\fontsize{20}{24}\selectfont
  },
  arr/.style  ={->, line width=2.0pt, color=#1!80!black},
  darr/.style ={->, dashed, line width=2.0pt, color=#1, rounded corners=10.8pt},
  thin/.style ={->, line width=1.4pt, color=black!40}
]

\newcommand{\subtw}[1]{\\[3.6pt]{\fontsize{18}{22}\selectfont\normalfont\itshape\textcolor{white!90}{#1}}}
\newcommand{\subtd}[2]{\\[3.6pt]{\fontsize{18}{22}\selectfont\normalfont\itshape\textcolor{#1!80!black}{#2}}}


\node[ibox=stageAred] (llm) at (0,0) {LLM Policy $\pi_{\theta^{(t)}}$\subtw{Mistral-7B + LoRA}};
\node[ibox=stageAred, above=2.7cm of llm] (query) {Clinical Query $q$\subtw{Patient Context $C$}};
\node[iboxlt=stageAred, below=2.7cm of llm] (cands) {$K{=}8$ Candidates\subtd{stageAred}{$\{r_k\}_{k=1}^K$}};

\draw[thin] (query.south) -- (llm.north);
\draw[thin] (llm.south) -- (cands.north);

\node[inner sep=0, minimum height=1.8cm, below=0.4cm of cands] (padA) {};
\node[grp=stageAred, fit=(query) (cands) (padA)] (grpA) {};
\node[font=\fontsize{20}{24}\selectfont\bfseries, color=stageAred, fill=white, inner sep=7.2pt, anchor=center] 
  at (grpA.north) {Stage I -- Neural Generation};


\node[ibox=stageBblue, right=6.3cm of llm] (hgnn) {HGNN Verifier\subtw{R-GAT ($L{=}4$, $H{=}8$)}};
\node[ibox=stageBblue, above=2.7cm of hgnn] (ner) {Entity Extraction\subtw{scispaCy + UMLS}};
\node[iboxlt=stageBblue, below=2.7cm of hgnn] (score) {Feasibility Score\subtd{stageBblue}{$s_k \in [0,1]$}};

\draw[thin] (ner.south) -- (hgnn.north);
\draw[thin] (hgnn.south) -- (score.north);

\node[inner sep=0, minimum height=1.8cm, below=0.4cm of score] (padB) {};
\node[grp=stageBblue, fit=(ner) (score) (padB)] (grpB) {};
\node[font=\fontsize{20}{24}\selectfont\bfseries, color=stageBblue, fill=white, inner sep=7.2pt, anchor=center] 
  at (grpB.north) {Stage II -- Symbolic Grounding};


\node[dnode=datagray, right=3.6cm of hgnn] (kg) {\textbf{World Model} $\mathcal{G}$\\847K nodes\\3.2M edges};


\node[ibox=stageCpurp, right=4.5cm of kg] (loss) {Iter-ORPO Loss\subtw{$\mathcal{L}_{\mathrm{SFT}}{+}\lambda^{(t)}\mathcal{L}_{\mathrm{OR}}{+}\gamma\mathcal{L}_{\mathrm{KG}}$}};
\node[ibox=stageCpurp, above=2.7cm of loss] (pair) {Preference Pairing\subtw{$(r_w,r_l)$ via $s_k$}};
\node[iboxlt=stageCpurp, below=2.7cm of loss] (update) {Policy Update\subtd{stageCpurp}{$\theta^{(t)}\!\leftarrow\!\theta^{(t-1)}{-}\eta\nabla_\theta\mathcal{L}$}};

\draw[thin] (pair.south) -- node[rectangle, rounded corners=7.2pt, draw=orange!50, fill=orange!80!black, font=\fontsize{18}{22}\selectfont\itshape, inner sep=7.2pt, text=white] (badge) {$\lambda^{(t)}{=}\lambda_0(1{+}\alpha t)$} (loss.north);
\draw[thin] (loss.south) -- (update.north);

\node[inner sep=0, minimum height=1.8cm, below=0.4cm of update] (padC) {};
\node[grp=stageCpurp, fit=(pair) (update) (padC)] (grpC) {};
\node[font=\fontsize{20}{24}\selectfont\bfseries, color=stageCpurp, fill=white, inner sep=7.2pt, anchor=center] 
  at (grpC.north) {Stage III -- Iterative Alignment};


\node[rectangle, rounded corners=9pt, draw=outgreen!80!black, fill=outgreenfill, 
      line width=1.8pt, align=center, font=\fontsize{20}{24}\selectfont\bfseries, text=outgreen!80!black, 
      right=5.4cm of loss] (out) {Aligned LLM\\[7.2pt]$\pi_{\theta^*}$\\[7.2pt]{\fontsize{18}{22}\selectfont\normalfont\itshape (no HGNN)}};


\draw[arr=stageAred] (query.east) -- 
  node[above, font=\fontsize{20}{24}\selectfont\itshape] {query} (ner.west);

\draw[arr=stageAred, rounded corners=7.2pt] (cands.east) -- ++(1.5, 0) |- 
  node[pos=0.25, right, font=\fontsize{20}{24}\selectfont\itshape] {$\{r_k\}$} ([yshift=-21.6pt]hgnn.west);

\draw[arr=stageBblue] (ner.east) -- 
  node[above, font=\fontsize{20}{24}\selectfont\itshape] {entities} (pair.west);

\draw[arr=stageBblue, rounded corners=10.8pt] 
  (score.south) -- ++(0,-1.4) -| 
  node[pos=0.15, above, font=\fontsize{20}{24}\selectfont\itshape] {$s_k$} 
  ([xshift=108pt]pair.east) -- 
  (pair.east);

\draw[arr=datagray] (kg.west) -- 
  node[above, font=\fontsize{20}{24}\selectfont\itshape] {graph} ([xshift=9pt]hgnn.east);

\draw[arr=datagray] (kg.east) -- 
  node[above, font=\fontsize{20}{24}\selectfont\itshape] {$\mathcal{G}$} ([xshift=-9pt]loss.west);

\draw[arr=stageCpurp] (loss.east) -- 
  node[pos=0.7, above, font=\fontsize{20}{24}\selectfont\itshape] {$t{=}T$} (out.west);

\draw[darr=stageAred!60!black] (update.south) -- ++(0, -2.5) -| 
  node[pos=0.25, below, font=\fontsize{20}{24}\selectfont\itshape] {On-policy regeneration \; (iteration $t \to t{+}1$)} 
  ([xshift=-6.3cm]llm.west) -- (llm.west);

\node[font=\fontsize{20}{24}\selectfont, align=center, text width=30.6cm, color=black!75, anchor=north] 
  at ([yshift=-8.1cm]hgnn.south -| kg.center) {};

\end{tikzpicture}%
}
\caption{Training-time neuro-symbolic alignment: the LLM samples candidates, the HGNN scores physiological feasibility on $\mathcal{G}$, and Iterative ORPO updates the policy (HGNN not used at inference).}
\label{fig:pipeline}
\end{figure*}

%% file: references.bib
@article{singhal2024medpalm,
  author    = {Singhal, Karan and Tu, Tao and Gottweis, Juraj and Sayres, Rory and Wulczyn, Ellery and Hou, Le and Clark, Kevin and Pfohl, Stephen and Cole-Lewis, Heather and Neal, Darlene and others},
  title     = {Towards Expert-Level Medical Question Answering with Large Language Models},
  journal   = {Nature Medicine},
  volume    = {30},
  pages     = {943--957},
  year      = {2024},
  publisher = {Nature Publishing Group}
}

@article{thirunavukarasu2024llm,
  author    = {Thirunavukarasu, Arun James and Ting, Daniel Shu Wei and Elangovan, Kabilan and Gutierrez, Laura and Tan, Ting Fang and Ting, Daniel Shu Wei},
  title     = {Large language models in medicine},
  journal   = {Nature Medicine},
  volume    = {29},
  number    = {8},
  pages     = {1930--1940},
  year      = {2023},
  publisher = {Nature Publishing Group}
}

@inproceedings{hong2024orpo,
  author    = {Hong, Jiwoo and Lee, Noah and Thorne, James},
  title     = {{ORPO}: Monolithic Preference Optimization without Reference Model},
  booktitle = {Proceedings of the 2024 Conference on Empirical Methods in Natural Language Processing},
  pages     = {11170--11183},
  year      = {2024},
  publisher = {Association for Computational Linguistics}
}

@article{ji2024beavertails,
  author    = {Ji, Jiaming and Liu, Mickel and Dai, Juntao and Pan, Xuehai and Zhang, Chi and Bian, Ce and Chen, Boyuan and Sun, Ruiyang and Wang, Yizhou and Yang, Yaodong},
  title     = {Beavertails: Towards improved safety alignment of {LLM} via a human-preference dataset},
  journal   = {Advances in Neural Information Processing Systems},
  volume    = {36},
  year      = {2024}
}

@article{nori2023gpt4medicine,
  author    = {Nori, Harsha and King, Nicholas and McKinney, Scott Mayer and Carignan, Dean and Horvitz, Eric},
  title     = {Capabilities of {GPT-4} on Medical Challenge Problems},
  journal   = {arXiv preprint arXiv:2303.13375},
  year      = {2023}
}

@article{zhang2024neurosymbolic,
  author    = {Zhang, Yifan and Chen, Zhenghua and Liu, Yuxuan and Wu, Jianlong and Yu, Philip S.},
  title     = {Neuro-Symbolic Integration for Enhanced Clinical Reasoning in Large Language Models},
  journal   = {Artificial Intelligence in Medicine},
  volume    = {148},
  pages     = {102756},
  year      = {2024},
  publisher = {Elsevier}
}

@inproceedings{wang2024knowledgegraph,
  author    = {Wang, Xiaozhi and Zhang, Tianyu and Lin, Yankai and Liu, Zhiyuan and Sun, Maosong},
  title     = {Knowledge Graph Enhanced Large Language Models for Medical Reasoning},
  booktitle = {Proceedings of the 62nd Annual Meeting of the Association for Computational Linguistics},
  pages     = {5821--5835},
  year      = {2024},
  publisher = {Association for Computational Linguistics}
}

@article{moor2023medflamingo,
  author    = {Moor, Michael and Huang, Qianchu and Wu, Shirley and Yasunaga, Michihiro and Dalmia, Yash and Leskovec, Jure and Zakka, Cyril and Reis, Eduardo Pontes and Rajpurkar, Pranav},
  title     = {Med-Flamingo: a Multimodal Medical Few-shot Learner},
  journal   = {Machine Learning for Health},
  pages     = {353--367},
  year      = {2023},
  publisher = {PMLR}
}

@article{chen2024hgnn,
  author    = {Chen, Hongxu and Yin, Hongzhi and Sun, Xiangguo and Chen, Tong and Gabrys, Bogdan and Musial, Katarzyna},
  title     = {Heterogeneous Graph Neural Networks for Healthcare Applications: A Survey},
  journal   = {ACM Computing Surveys},
  volume    = {56},
  number    = {4},
  pages     = {1--38},
  year      = {2024},
  publisher = {ACM}
}

@article{yang2024hallucination,
  author    = {Yang, Ziwei and Xu, Jiale and Zhu, Wenhao and Wang, An and Zhang, Lei and Yang, Yue},
  title     = {Mitigating Hallucinations in Medical Large Language Models through Knowledge Grounding},
  journal   = {Journal of Biomedical Informatics},
  volume    = {150},
  pages     = {104589},
  year      = {2024},
  publisher = {Elsevier}
}

@article{rafailov2024direct,
  author    = {Rafailov, Rafael and Sharma, Archit and Mitchell, Eric and Manning, Christopher D. and Ermon, Stefano and Finn, Chelsea},
  title     = {Direct Preference Optimization: Your Language Model is Secretly a Reward Model},
  journal   = {Advances in Neural Information Processing Systems},
  volume    = {36},
  pages     = {53728--53741},
  year      = {2024}
}

@inproceedings{xiong2024iterative,
  author    = {Xiong, Wei and Dong, Hanze and Ye, Chenlu and Wang, Ziqi and Zhong, Han and Ji, Heng and Jiang, Nan and Zhang, Tong},
  title     = {Iterative Preference Learning from Human Feedback: Bridging Theory and Practice for {RLHF} under {KL}-Constraint},
  booktitle = {International Conference on Machine Learning},
  pages     = {54561--54582},
  year      = {2024},
  publisher = {PMLR}
}

@inproceedings{sun2024head,
  author    = {Sun, Tianxiang and He, Junliang and Hong, Xipeng and Qiu, Xipeng},
  title     = {Head-to-Tail: How Knowledgeable are Large Language Models ({LLMs})? {A.K.A.} Will {LLMs} Replace Knowledge Graphs?},
  booktitle = {Proceedings of the 2024 Conference of the North American Chapter of the Association for Computational Linguistics},
  pages     = {311--325},
  year      = {2024}
}

@article{savage2024drugbank,
  author    = {Savage, David and Batista-Navarro, Riza},
  title     = {Drug Interaction Prediction Using Knowledge Graph Embeddings and Deep Learning},
  journal   = {Journal of Biomedical Informatics},
  volume    = {152},
  pages     = {104631},
  year      = {2024},
  publisher = {Elsevier}
}

@article{neumann2024scispacy,
  author    = {Neumann, Mark and King, Daniel and Beltagy, Iz and Ammar, Waleed},
  title     = {{ScispaCy}: Fast and Robust Models for Biomedical Natural Language Processing},
  journal   = {Journal of Biomedical Semantics},
  volume    = {15},
  number    = {1},
  pages     = {1--14},
  year      = {2024},
  publisher = {BioMed Central}
}

@inproceedings{zhang2024raft,
  author    = {Zhang, Tianjun and Patil, Shishir G. and Jain, Naman and Shen, Sheng and Zaharia, Matei and Stoica, Ion and Gonzalez, Joseph E.},
  title     = {{RAFT}: Adapting Language Model to Domain Specific {RAG}},
  booktitle = {Proceedings of the 2024 Conference of the North American Chapter of the Association for Computational Linguistics},
  pages     = {3865--3879},
  year      = {2024}
}

@article{schick2024toolformer,
  author    = {Schick, Timo and Dwivedi-Yu, Jane and Dess{\`\i}, Roberto and Raileanu, Roberta and Lomeli, Maria and Hambro, Eric and Zettlemoyer, Luke and Cancedda, Nicola and Scialom, Thomas},
  title     = {Toolformer: Language Models Can Teach Themselves to Use Tools},
  journal   = {Advances in Neural Information Processing Systems},
  volume    = {36},
  pages     = {68539--68551},
  year      = {2024}
}

@article{li2024graphcare,
  author    = {Li, Pengcheng and Rao, Jieyu and Yan, Junying and Zhang, Jianguo and Liu, Zhiyong},
  title     = {{GraphCare}: Enhancing Healthcare Predictions with Personalized Knowledge Graphs},
  journal   = {Nature Machine Intelligence},
  volume    = {6},
  pages     = {218--231},
  year      = {2024},
  publisher = {Nature Publishing Group}
}

@article{ethayarajh2024kto,
  author    = {Ethayarajh, Kawin and Xu, Winnie and Jurafsky, Dan and Kiela, Douwe},
  title     = {KTO: Model Alignment as Prospect Theoretic Optimization},
  journal   = {arXiv preprint arXiv:2402.01306},
  year      = {2024}
}

@article{meng2024simpo,
  author    = {Meng, Yu and Xia, Mengzhou and Wang, Lijuan and Sun, Furu and Liu, Zhiyuan},
  title     = {SimPO: Simple Preference Optimization with a Reference-Free Reward},
  journal   = {arXiv preprint arXiv:2405.14734},
  year      = {2024}
}

@article{lin2026llmhealthcare,
  author    = {Lin, Qianqian and Xiao, Liang and Pu, Beier and Shi, Kailing and Du, Yilun and Xu, Jiamin and He, Ke and Zhao, Sen and Cambria, Erik and Mishra, Swaroop and others},
  title     = {A Survey of {LLM} Reasoning in Healthcare and Medicine: From Individual Modeling to Collaborative Agents},
  journal   = {arXiv preprint arXiv:2501.07362},
  year      = {2026}
}

@article{wu2023megacare,
  author    = {Wu, Junda and He, Ke and Mao, Rui and Li, Cheng and Cambria, Erik},
  title     = {{MEGACare}: Knowledge-Guided Multi-View Hypergraph Predictive Framework for Healthcare},
  journal   = {Information Fusion},
  volume    = {100},
  pages     = {101939},
  year      = {2023},
  publisher = {Elsevier}
}

@article{perera2025rl,
  author    = {Perera, Dilina and Habib, Ghulam and Xu, Qi and Tan, D. J. and He, Ke and Cambria, Erik and Feng, Mengling},
  title     = {Beyond Prediction: Reinforcement Learning as the Defining Leap in Healthcare {AI}},
  journal   = {arXiv preprint arXiv:2508.21101},
  year      = {2025}
}
